\documentclass[10pt,journal,compsoc]{IEEEtran}
\ifCLASSOPTIONcompsoc
  \usepackage[nocompress]{cite}
\else
  \usepackage{cite}
\fi
\ifCLASSINFOpdf
\else
\fi

\usepackage{cite}
\usepackage{graphicx}
\usepackage{amsmath}
\usepackage{array}
\usepackage{amssymb}
\usepackage{booktabs}
\usepackage{bbding}
\usepackage{subfigure}
\usepackage{textcomp}
\usepackage{fixltx2e}
\usepackage{url}
\usepackage{multirow}
\usepackage{color}
\def\ie{{\em i.e.}}
\def\eg{{\em e.g.}}
\def\etal{{\em et al.}}
\begin{document}
\title{RefineFace: Refinement Neural Network for High Performance Face Detection}
\author{Shifeng~Zhang$^*$,
        Cheng~Chi$^*$,
        Zhen~Lei$^\dag$,~\IEEEmembership{Senior Member,~IEEE,}
        and~Stan~Z.~Li,~\IEEEmembership{Fellow,~IEEE}
\thanks{Shifeng Zhang, Zhen Lei and Stan Z. Li are with the Center for Biometrics and Security Research (CBSR), National Laboratory of Pattern Recognition (NLPR), Institute of Automation Chinese Academy of Sciences (CASIA) and University of Chinese Academy of Sciences (UCAS), Beijing, China (e-mail: \{shifeng.zhang, zlei, szli\}@nlpr.ia.ac.cn).}
\thanks{Cheng Chi is with the Institute of Electronics Chinese Academy of Sciences (IECAS) and University of Chinese Academy of Sciences (UCAS), Beijing, China (e-mail: chicheng15@mails.ucas.ac.cn).}
\thanks{$^*$Equal contribution. \quad $^\dag$ Corresponding author.}
}

\markboth{Journal of xx,~Vol.~xx, No.~x, July~2019}%
{Zhang \MakeLowercase{\textit{et al.}}: RefineFace: Refinement Neural Network for High Performance Face Detection}

\IEEEtitleabstractindextext{
\begin{abstract}
Face detection has achieved significant progress in recent years. However, high performance face detection still remains a very challenging problem, especially when there exists many tiny faces. In this paper, we present a single-shot refinement face detector namely RefineFace to achieve high performance. Specifically, it consists of five modules: Selective Two-step Regression (STR), Selective Two-step Classification (STC), Scale-aware Margin Loss (SML), Feature Supervision Module (FSM) and Receptive Field Enhancement (RFE). To enhance the regression ability for high location accuracy, STR coarsely adjusts locations and sizes of anchors from high level detection layers to provide better initialization for subsequent regressor. To improve the classification ability for high recall efficiency, STC first filters out most simple negatives from low level detection layers to reduce search space for subsequent classifier, then SML is applied to better distinguish faces from background at various scales and FSM is introduced to let the backbone learn more discriminative features for classification. Besides, RFE is presented to provide more diverse receptive field to better capture faces in some extreme poses. Extensive experiments conducted on WIDER FACE, AFW, PASCAL Face, FDDB, MAFA demonstrate that our method achieves state-of-the-art results and runs at $37.3$ FPS with ResNet-18 for VGA-resolution images.
\end{abstract}

}

\maketitle

\IEEEdisplaynontitleabstractindextext
\IEEEpeerreviewmaketitle

\section{Introduction}
\IEEEPARstart{F}{ace} detection is a long-standing problem in computer vision with many applications, such as face alignment, face analysis, face recognition and face tracking. Given an image, the goal of face detection is to determine whether there are any faces, and if any, return the bounding box of each face. To detect faces efficiently and accurately, different detection pipelines have been designed after the pioneering work of Viola-Jones~\cite{DBLP:journals/ijcv/ViolaJ04}. Among them, the single-shot anchor-based approach~\cite{DBLP:conf/aaai/abs-1809-02693, li2018dsfd, DBLP:conf/iccv/NajibiSCD17, tang2018pyramidbox, DBLP:conf/iccv/abs-1708-05237} is the dominant method. It performs face detection based on regular and dense anchors over various locations, scales and aspect ratios. In this framework, the face detection task is decomposed into two sub-tasks: the binary classification and the bounding box regression. The former one aims to classify the preset anchor boxes into face and background, and the latter one is to regress those detected faces to more accurate locations.

\begin{figure}[t]
\centering
\includegraphics[width=1.0\linewidth]{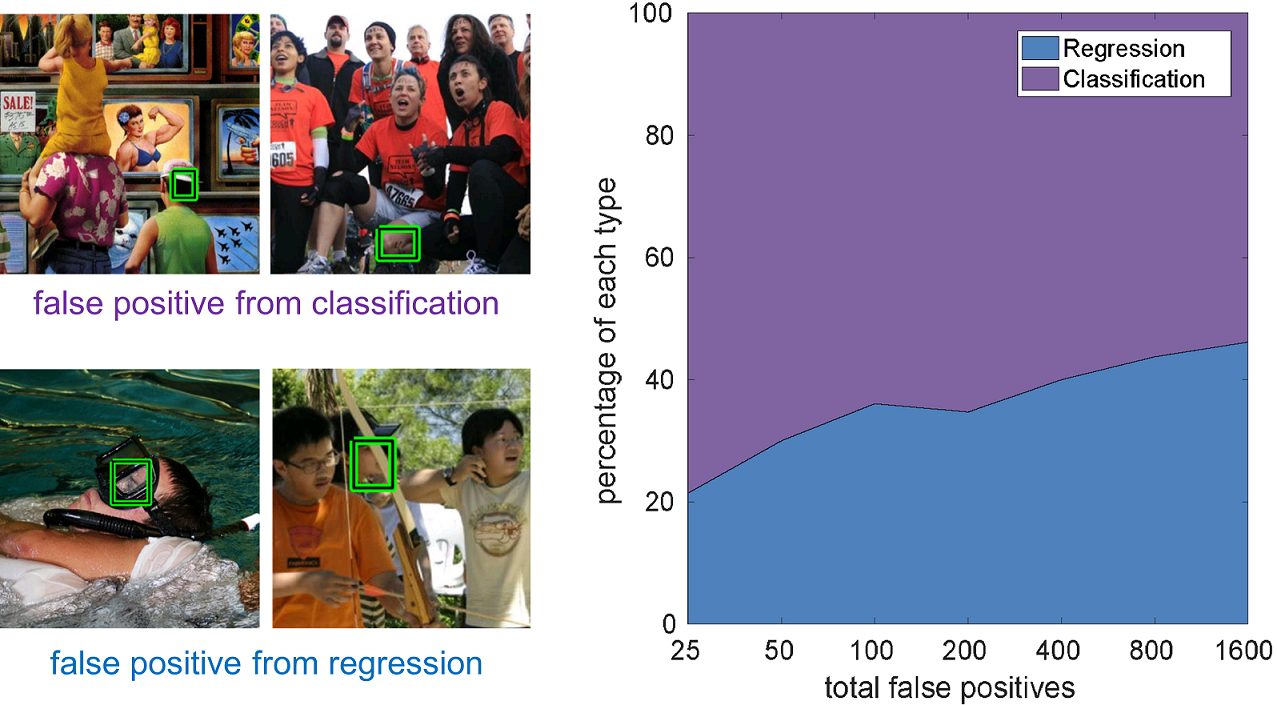}
\caption{Illustration of false positives of our baseline face detector on the WIDER FACE validation Hard subset. Left: Example of two error types of false positives. Right: Distribution of two error types of false positives.}
\label{fig:err_dist}
\end{figure}

With the development of deep convolutional neural networks (CNNs), single-shot anchor-based face detectors have been thoroughly studied and great progress has been made in recent years. In particular, on the very challenging face detection dataset WIDER FACE~\cite{DBLP:conf/cvpr/YangLLT16}, the average precision (AP) on its Hard subset has been improved from $40.0\%$ to $90.0\%$ by recent algorithms~\cite{DBLP:conf/aaai/abs-1809-02693, li2018dsfd, DBLP:conf/iccv/NajibiSCD17, tang2018pyramidbox, DBLP:conf/iccv/abs-1708-05237} over the past three years. For now, it has become a challenging problem to further improve these single-shot face detectors' performance, especially when there exists many tiny faces. In our opinion, there remains room for improvement in two aspects: a) location accuracy: accuracy of the bounding box location needs to be improved, \ie, boosting the regression ability; b) recall efficiency: more faces need to be recalled with less false positives, \ie, enhancing the classification ability. To embody these two aspects still can be improved, we utilize the detection analysis tool\footnote{\url{http://web.engr.illinois.edu/~dhoiem/projects/detectionAnalysis}} to analyze the error distribution of our baseline face detector RetinaNet~\cite{DBLP:conf/iccv/LinPRK17} on the WIDER FACE validation Hard subset. As shown in Figure~\ref{fig:err_dist}, there are two error types of false positives from face detectors: the Regression (LOC) error indicates that a face is detected with a misaligned localization, and the Classification (CLS) error means that a background region is mistakenly detected as a face. Apart from false positives, false negatives indicate that a face fails to be detected, which also belong to the CLS error. These two error modes are elaborated as follows.

\begin{figure}[t]
\centering
\subfigure[Location Accuracy]{
\label{fig:la}
\includegraphics[width=0.4825\linewidth]{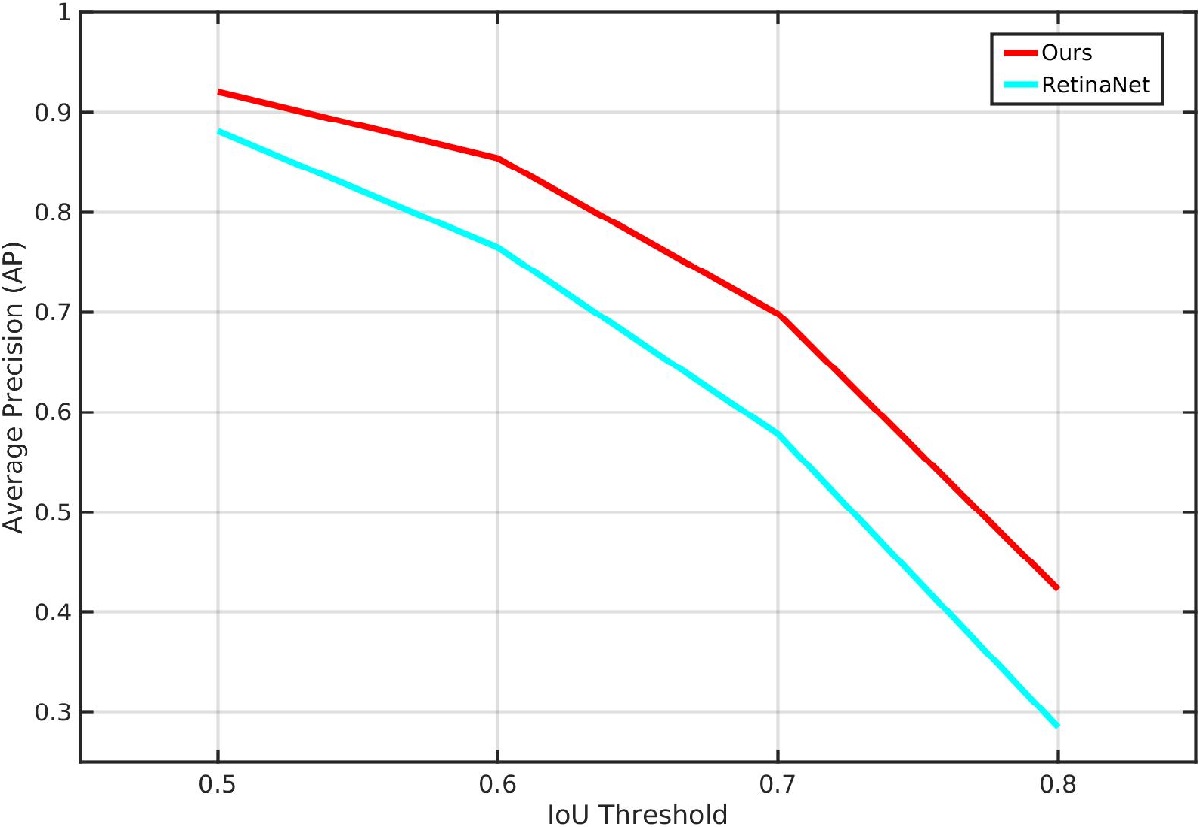}}
\subfigure[Recall Efficiency]{
\label{fig:re}
\includegraphics[width=0.4825\linewidth]{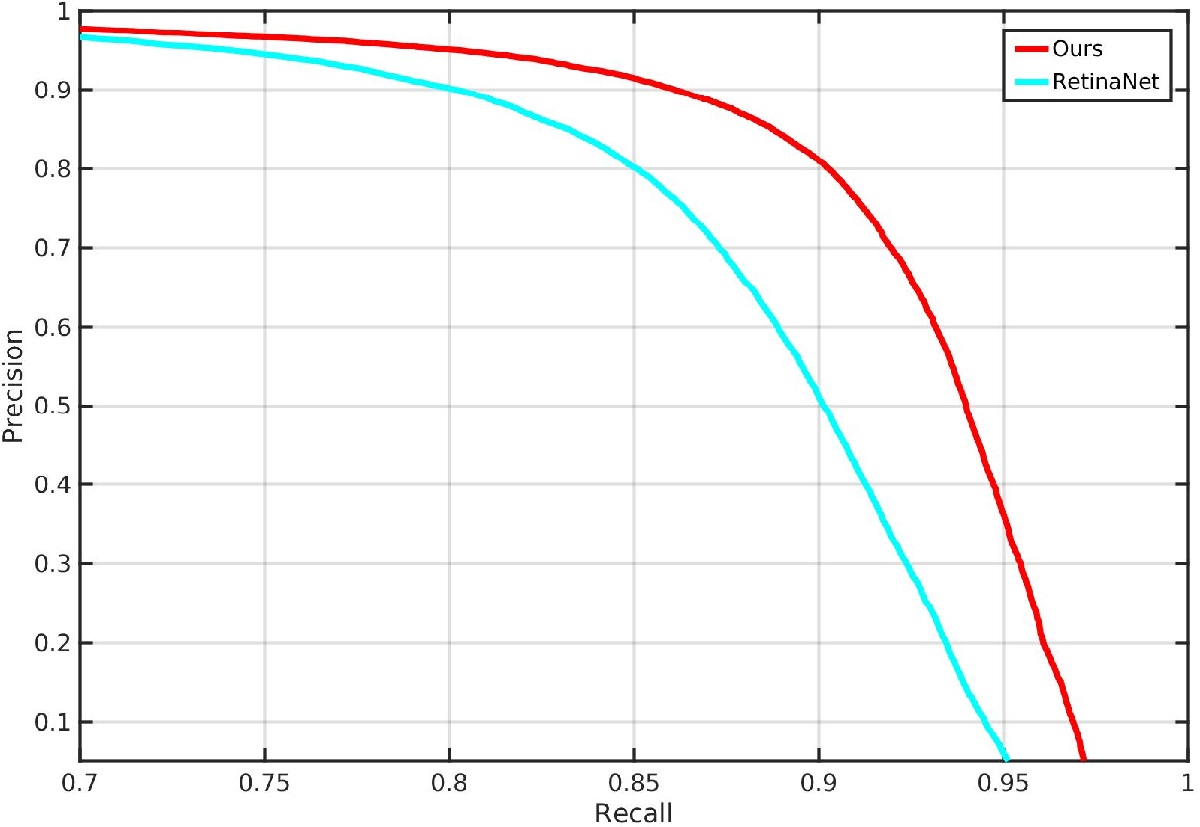}}
\caption{(a) As the IoU threshold increases, the AP of RetinaNet drops dramatically. Our method improves its location accuracy by boosting the regression ability. (b) RetinaNet produces about $50\%$ false positives when the recall rate is $90\%$ and it also misses about $5\%$ faces. Our method improves its recall efficiency by enhancing the classification ability.}
\label{fig:rela}
\end{figure}

The LOC error is triggered by the lack of strong regression ability and its manifestation is plenty of false detections with inaccurate localization. If the regression ability of the face detector can be enhanced, these false positves from the LOC error will be reduced. Actually, the location accuracy in the face detection task has attracted much more attention of researchers in recent years. Although current evaluation criteria of most face detection datasets~\cite{DBLP:conf/cvpr/ZhuR12,DBLP:journals/ivc/YanZLL14,fddbTech} do not focus on the location accuracy, the WIDER FACE Challenge\footnote{http://wider-challenge.org} adopts MS COCO~\cite{DBLP:conf/eccv/LinMBHPRDZ14} evaluation criterion, which puts more emphasis on bounding box location accuracy. To visualize the location accuracy issue, we use different IoU thresholds to evaluate our baselinse face detector RetinaNet~\cite{DBLP:conf/iccv/LinPRK17} on the WIDER FACE dataset. As shown in Figure~\ref{fig:la}, as the IoU threshold increases, the AP drops dramatically, indicating that the accuracy of the bounding box location needs to be improved. To this end, Gidaris et al.~\cite{DBLP:conf/iccv/GidarisK15} propose iterative regression during inference to improve the accuracy. Cascade R-CNN~\cite{DBLP:journals/corr/abs-1712-00726} addresses this issue by cascading R-CNN with different IoU thresholds. RefineDet~\cite{DBLP:conf/cvpr/ZhangWBLL18} applies two-step regression to single-shot detector. However, blindly adding multi-step regression to the face detection task is often counterproductive, which needs more exploration.

The CLS error is caused by the non-robust classification ability and its manifestation is lots of false alarms and missing faces. Specifically, the average precision (AP) of current face detection algorithms is already very high, but the recall efficiency is not high enough. As shown in Figure~\ref{fig:re} of our baseline, its precision is only about $50\%$ (half of detections are false alarms) when the recall rate is equal to $90\%$, and its highest recall rate is only about $95\%$ (the remaining $5\%$ faces are still not detected). Reflected on the shape of the Precision-Recall curve, it has not extended far enough to the right as well as not steep enough. In our opinion, the reasons behind the recall efficiency issue in the CLS error of single-shot detectors are 1) class imbalance: plenty of small anchors need to be tiled to detect tiny faces, causing the extreme class imbalance problem; 2) scale problem: different scales of anchors have different degrees of classification difficulty, and the classification of smaller anchors are more difficult; 3) feature misalignment: different anchors at the same location are classified based on the same misaligned features. They are the culprit leading to the CLS error. If we can improve the recall efficiency via enhancing the classification ability, more faces can be correctly detected from the complex background, making fewer CLS errors and higher average precision (AP). Therefore, it is worth further study to improve the classification ability of the face detector.

To reduce the aforementioned two errors, we propose five improvements to enhance the regression and classification ability of the high performance face detector and present a new state-of-the-art method namely RefineFace. Specifically, to improve the regression ability, we apply the STR to coarsely adjust the locations and sizes of anchors from high level detection layers to provide better initialization for the subsequent regressor. To enhance the classification ability, we first use the STC to filter out most simple negatives from low level detection layers to reduce the search space for the subsequent classifier, then employ the SML to better distinguish faces from background at various scales and the FSM to let the backbone network learn more discriminative features for classification. Besides, we design the RFE to provide more diverse receptive field to better capture faces in some extreme poses. We conduct extensive experiments on the WIDER FACE, AFW, PASCAL Face, FDDB and MAFA benchmark datasets and achieve the state-of-the-art results with $37.3$ FPS for the VGA-resolution image. The main contributions of this paper can be summarized below.

\begin{itemize}
\item Designing a STR module to coarsely adjust the locations and sizes of anchors from high level layers to provide better initialization for the subsequent regressor.
\item Presenting a STC module to filter out most simple negative samples from low level layers to reduce the classification search space.
\item Introducing a SML module to better distinguish faces from background across different scales.
\item Proposing a FSM module to learn more discriminative features for the classification task.
\item Constructing a RFE module to provide more diverse receptive fields for detecting extreme-pose faces.
\item Achieving state-of-the-art performances on AFW, PASCAL face, FDDB, MAFA and WIDER FACE datasets.
\end{itemize}

\begin{figure*}[ht!]
\centering
\includegraphics[width=0.92\linewidth]{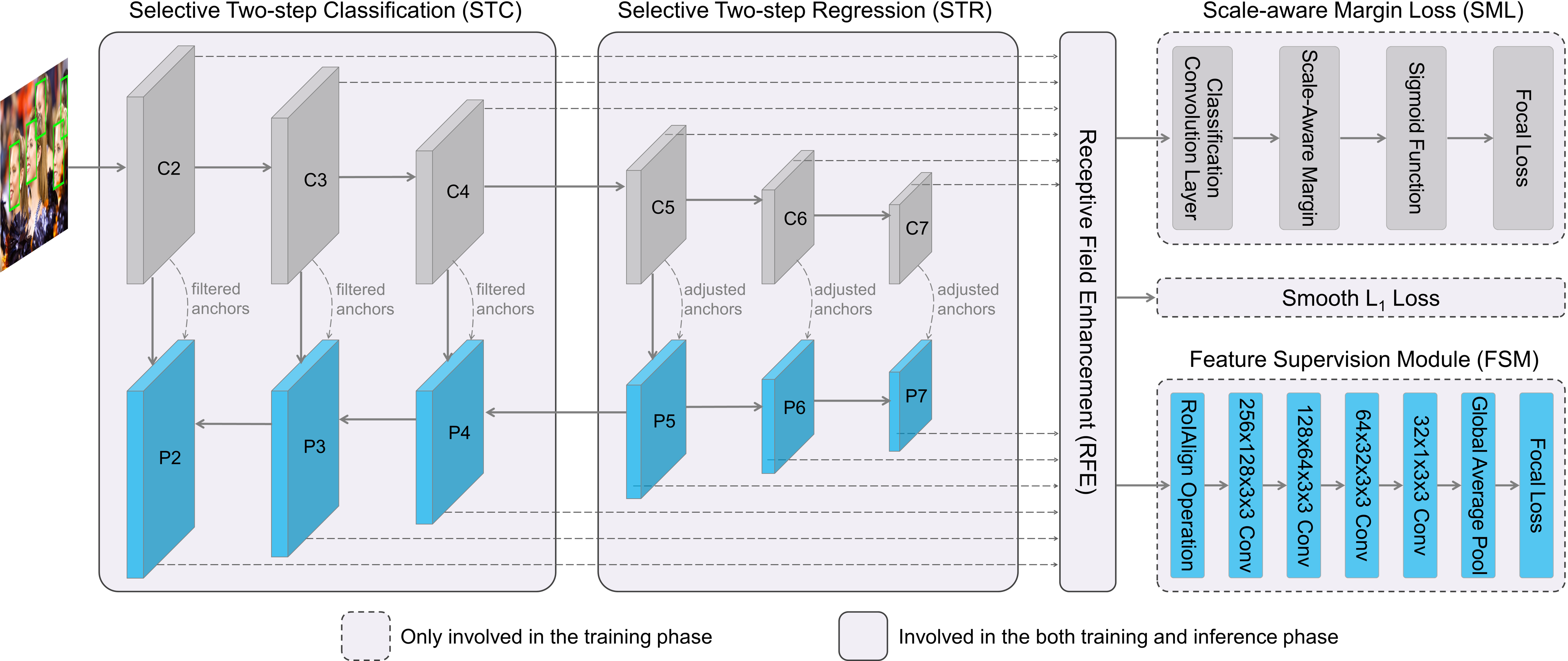}
\caption{Structure of RefineFace. It is based on RetinaNet with five proposed modules. Among them, SML and FAM are only involved in training without any overhead in inference, while STC, STR and RFE introduce a small amount of overhead.}
\label{fig:framework}
\end{figure*}

Preliminary results of this work have been published in \cite{DBLP:conf/aaai/abs-1809-02693}. The current work has been improved and extended from the conference version in several important aspects. (1) We introduce a Scale-aware Margin Loss (SML) function to better distinguish faces from the complex background across different scales. (2) We design a Feature Supervision Module (FSM) to learn more discriminative features for classification. (3) We noticeably improve the accuracy of the detector in our previous work without introducing any additional overhead during the inference phase. (4) All sections are rewritten with more details, more references and more experiments to have a more elaborate presentation.

\section{Related Work} \label{2}
Face detection has attracted much attention these years for its wide practical applications. The pioneering work of Viola and Jones~\cite{DBLP:journals/ijcv/ViolaJ04} uses AdaBoost with Haar features to train a cascaded face detector and inspires several different approaches afterwards~\cite{DBLP:journals/ijcv/BrubakerWSMR08, DBLP:journals/pami/LiaoJL16, DBLP:conf/iccv/PhamC07}. Besides, the Deformable Part Model (DPM)~\cite{DBLP:journals/pami/FelzenszwalbGMR10} is another popular framework in traditional face detection~\cite{DBLP:conf/eccv/MathiasBPG14,DBLP:conf/cvpr/YanLWL14,DBLP:conf/cvpr/ZhuR12}. However, the aforementioned methods are unreliable in complex scenarios because of non-robust hand-crafted features and classifiers.

In recent years, face detection has been dominated by CNN-based methods. Li~\etal~\cite{DBLP:conf/cvpr/LiLSBH15} achieve promising accuracy and efficiency by separately training a series of CNN models and following work~\cite{DBLP:conf/cvpr/QinYLH16} realizes end-to-end optimization. Yang~\etal~\cite{DBLP:conf/iccv/YangLLT15} detect faces under severe occlusion and unconstrained pose variations via scoring facial parts responses according to their spatial structure and arrangement. Ohn-Bar~\etal~\cite{DBLP:conf/icpr/Ohn-BarT16a} utilize the boosted decision tree classifier to detect faces. Yu~\etal~\cite{DBLP:conf/mm/YuJWCH16} propose an IoU loss to directly regress the bounding box that is robust to objects of varied shapes and scales. Zhang~\etal~\cite{DBLP:journals/spl/ZhangZLQ16} use multi-task cascaded CNNs to jointly detect faces and landmarks. Yang~\etal~\cite{DBLP:journals/corr/YangXLT17} apply a specialized set of CNNs with different structures to detect different scales of faces. Hu~\etal~\cite{DBLP:conf/cvpr/HuR17} train some separate detectors for different scales to find tiny faces. Zhu~\etal~\cite{zhu2018seeing} introduce an EMO score to evaluate the quality of anchor setting. Hao~\etal~\cite{DBLP:conf/cvpr/HaoLQYLH17} develop a scale proposal stage to guide the zoom-in and zoom-out of image to detect normalized face. Song~\etal~\cite{DBLP:conf/cvpr/Song0JWYL18} propose a scale estimation and spatial attention proposal module to pay attention to some specific scales and valid locations in the image pyramid. Shi~\etal~\cite{DBLP:conf/cvpr/ShiSKWC18} detect rotated faces in a coarse-to-fine manner under a cascade-style structure. Bai~\etal~\cite{DBLP:conf/cvpr/BaiZDG18} utilize GAN~\cite{goodfellow2014generative} to detect blurry small faces via generating clear super-resolution ones.

In addition, generic object detection algorithms have inspired many face detection methods. CMS-RCNN~\cite{DBLP:journals/corr/ZhuZLS16} integrates contextual reasoning into Faster R-CNN~\cite{DBLP:journals/pami/RenHG017} to help reduce the overall detection errors. Face R-CNN~\cite{DBLP:journals/corr/WangLJW17}, Face R-FCN~\cite{DBLP:journals/corr/abs-1709-05256} and FDNet~\cite{DBLP:journals/corr/abs-1802-02142} apply Faster R-CNN~\cite{DBLP:journals/pami/RenHG017} and R-FCN~\cite{DBLP:conf/nips/DaiLHS16} with some specific strategies to perform face detection. FaceBoxes~\cite{DBLP:conf/ijcb/abs-1708-05234} designs a CPU real-time face detector based on SSD~\cite{DBLP:conf/eccv/LiuAESRFB16}. S$^{3}$FD~\cite{DBLP:conf/iccv/abs-1708-05237} introduces some specific strategies onto SSD~\cite{DBLP:conf/eccv/LiuAESRFB16} to alleviate the matching problem of small faces. SSH~\cite{DBLP:conf/iccv/NajibiSCD17} adds large filters on each prediction head to merge the context information. PyramidBox~\cite{tang2018pyramidbox} takes advantage of the information around human faces to improve detection performance. FAN~\cite{DBLP:journals/corr/abs-1711-07246} utilizes the anchor-level attention mechanism onto RetinaNet~\cite{DBLP:conf/iccv/LinPRK17} to detect the occluded faces. FANet~\cite{DBLP:journals/corr/abs-1712-00721} aggregates higher-level features like FPN~\cite{DBLP:conf/cvpr/LinDGHHB17} to augment lower-level features at marginal extra computation cost. DFS~\cite{DBLP:journals/corr/abs-1811-08557} introduces a more effective feature fusion pyramid and a more efficient segmentation branch to handle hard faces. DSFD~\cite{li2018dsfd} inherits the architecture of SSD~\cite{DBLP:conf/eccv/LiuAESRFB16} and introduces a feature enhance module to extend the single shot detector to dual shot detector. SRN~\cite{DBLP:conf/aaai/abs-1809-02693} combines the multi-step detection in RefineDet~\cite{DBLP:conf/cvpr/ZhangWBLL18} and the focal loss in RetinaNet~\cite{DBLP:conf/iccv/LinPRK17} to perform efficient and accurate face detection. VIM-FD~\cite{DBLP:journals/corr/abs-1901-02350} and ISRN~\cite{DBLP:journals/corr/abs-1901-06651} improve SRN~\cite{DBLP:conf/aaai/abs-1809-02693} with data augmentation, attention mechanism and training from scratch.

\section{RefineFace}\label{3}

The overall architecture of RefineFace is shown in Figure~\ref{fig:framework}. We adopt ResNet~\cite{DBLP:conf/cvpr/HeZRS16} with 6-level feature pyramid structure as backbone for RefineFace. The feature maps extracted from those four residual blocks are denoted as C2, C3, C4, and C5, respectively. C6 and C7 are extracted by two simple down-sample $3\times3$ convolution layers after C5. The lateral structure between the bottom-up and the top-down pathways is the same as~\cite{DBLP:conf/cvpr/LinDGHHB17}. P2, P3, P4, and P5 are the feature maps extracted from lateral connections, corresponding to C2, C3, C4, and C5 that are respectively of the same spatial sizes, while P6 and P7 are down-sampled by two $3\times3$ convolution layers after P5. The proposed RefineFace is based on our baseline face detector RetinaNet with five newly proposed modules:
\begin{itemize}
\item STR: It selects C5, C6, C7, P5, P6, and P7 to conduct two-step regression.
\item STC: It selects C2, C3, C4, P2, P3, and P4 to perform two-step classification.
\item SML: It adds the scale-aware margin to the classification loss to better distinguish faces from background across different scales.
\item FSM: It contains one RoIAlign layer, four $3\times3$ convolution layers, one global average pooling layer and the Focal loss to let backbone learn more discriminative features for the classification task.
\item RFE: It enriches the receptive field of features used to predict the classification and location of objects.
\end{itemize}

Without the above proposed five modules, it is our baseline face detector, consisting of C2-C5 and P2-P7 with the Focal loss and the smooth L$_1$ loss. As shown in Table~\ref{tab:ablation}, we aim to boost its regression and classification ability to obtain a new state-of-the-art method.

\subsection{Selective Two-step Regression}
Single-shot detectors conduct only one regression operation to get final detections from anchors. Comparing to two-stage detectors with multi-step regression, single-shot detectors with one-step regression lack of strong regression ability. This inadequacy causes lots of inaccurate detection results, which will be considered as false positives, especially under MS COCO-style evaluation standard. To this end, using cascade structure~\cite{DBLP:conf/cvpr/ZhangWBLL18,DBLP:journals/corr/abs-1712-00726} to conduct multi-step regression is an effective method to improve the regression ability for accurate detection bounding boxes.

However, blindly adding multi-step regression to the face detection task is often counterproductive. Specifically, experimental results in Table~\ref{tab:str_per_level} indicate that applying two-step regression in the three lower pyramid levels impairs the performance. The reasons behind this phenomenon are twofold: 1) the three lower pyramid levels are associated with plenty of small anchors to detect small faces. These small faces are characterized by very coarse feature representations, so it is difficult for these small anchors to perform two-step regression; 2) in the training phase, if we let the network pay too much attention to the difficult regression task on the low pyramid levels, it will cause larger regression loss and hinder the more important classification task. In contrast, the three higher pyramid levels are associated with a small number of large anchors to detect large faces with detailed features, conducting two-step regression in these three levels is feasible and will improve the performance as shown in Table~\ref{tab:str_per_level}.

\begin{figure}[t]
\centering
\subfigure[Adjusted Anchor]{
\label{fig:aa}
\includegraphics[width=0.45\linewidth]{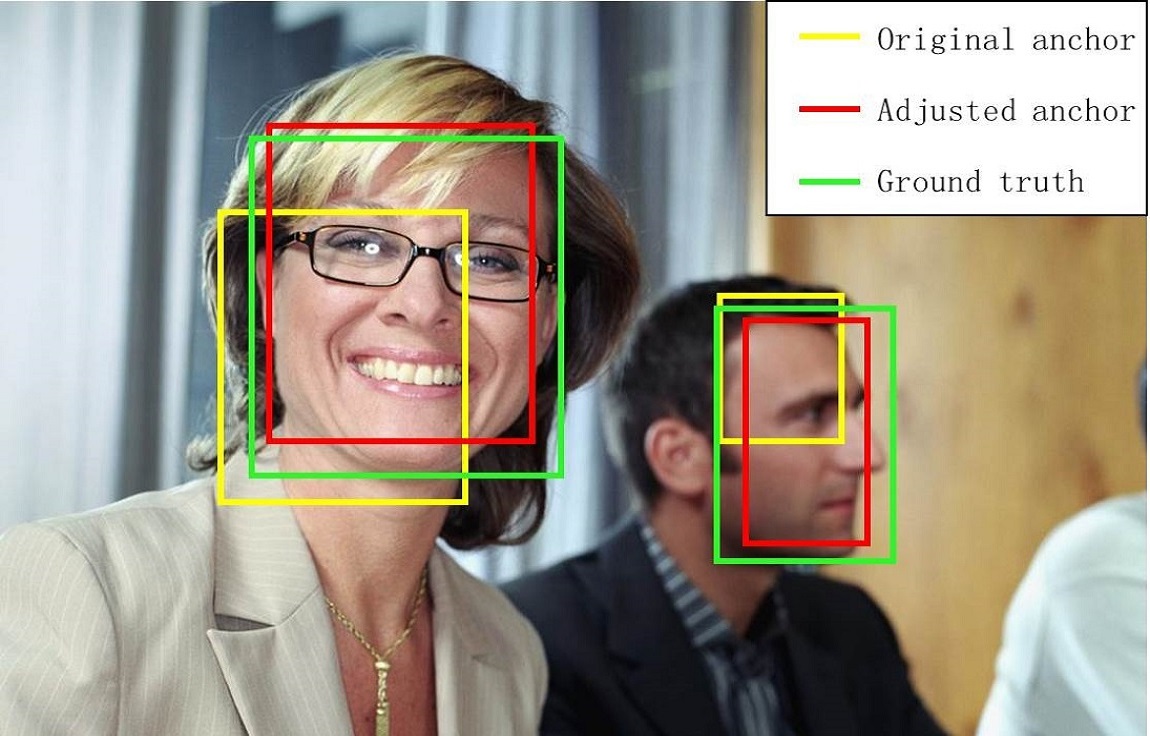}}
\subfigure[Effect on Class Imbalance]{
\label{fig:sc}
\includegraphics[width=0.45\linewidth]{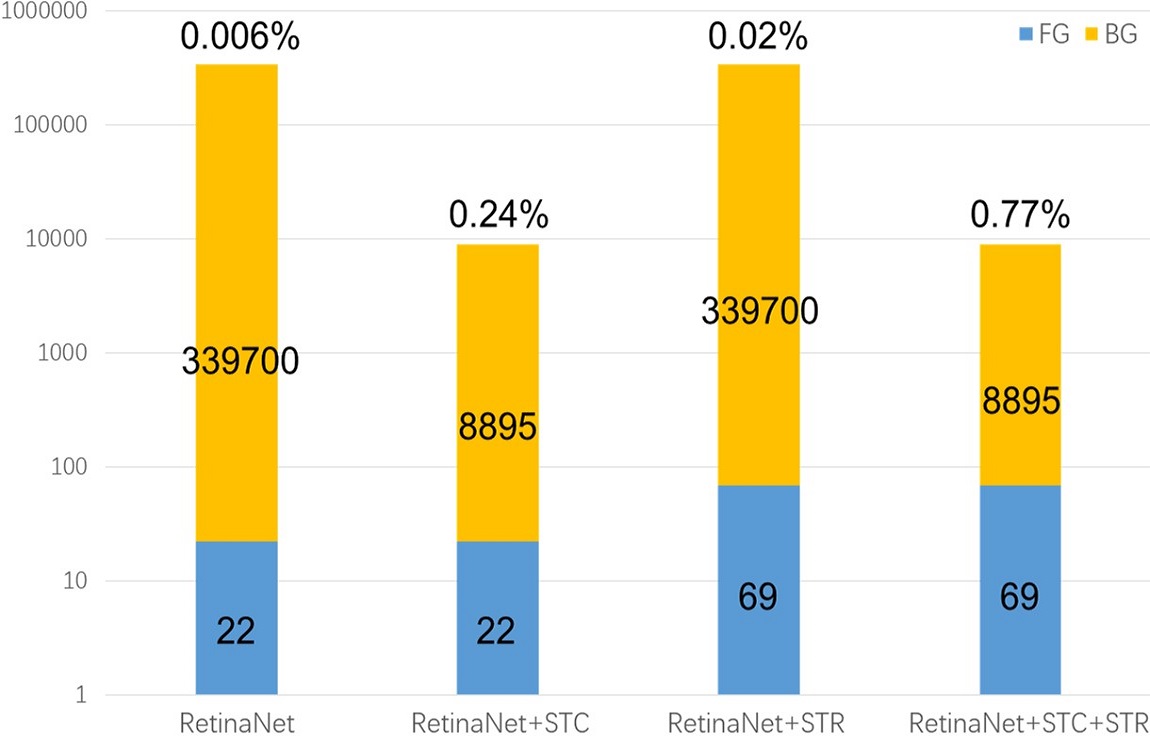}}
\caption{(a) STR provides better initialization for the subsequent regressor. (b) STC increases the positives/negatives ratio by about $38$ times.}
\label{fig:scaa}
\end{figure}

Based on the above analyses, we selectively perform two-step regression on the three higher pyramid levels. As shown in Figure~\ref{fig:aa}, the STR coarsely adjusts the locations and sizes of anchors from high levels of detection layers to provide better initialization for the subsequent regressor, which can enhance the regression ability to regress more accurate locations of bounding boxes. The loss function of STR consists of two parts, which is shown below:

\begin{small}
\begin{equation}
{\cal L}_\text{STR}=\frac{1}{N_{\text{s}_1}} \sum_{i\in \Psi}[l_i^\ast\!\!=\!\!1]{\cal L}_{\text{r}}(x_i,g_i^\ast) + \frac{1}{N_{\text{s}_2}} \sum_{i\in \Phi}[l_i^\ast\!\!=\!\!1]{\cal L}_{\text{r}}(t_i,g_i^\ast),
\label{eqn1}
\end{equation}
\end{small}
{\flushleft where $i$ is the index of anchor in a mini-batch, $l_i^\ast$ and $g_i^\ast$ are the ground truth class label and size of anchor $i$, $x_i$ is the refined coordinates of anchor $i$ in the first step, $t_i$ is the coordinates of the bounding box in the second step, $N_{\text{$s_1$}}$ and $N_{\text{$s_2$}}$ are the numbers of positive anchors in the first and second steps, $\Psi$ represents a collection of samples selected for two-step regression, and $\Phi$ represents a sample set in the second step. Similar to Faster R-CNN \cite{DBLP:journals/pami/RenHG017}, we use the smooth L$_1$ loss as the regression loss $L_{\text{r}}$. The Iverson bracket indicator function $[l_i^\ast=1]$ outputs $1$ when the condition is true, \ie, $l_i^\ast=1$ (the anchor is not the negative), and $0$ otherwise. Hence $[l_i^\ast=1]{\cal L}_{\text{r}}$ indicates that the regression loss is ignored for negative anchors.}

\subsection{Selective Two-step Classification}
It is necessary for single-shot anchor-based face detectors to tile plenty of anchors over the image to detect faces of various scales, which causes the extreme class imbalance between the positive and negative samples. For example, in our RefineFace structure with the $1024\times1024$ input resolution, if we tile $2$ anchors at each anchor point, the total number of samples will reach $300k$. Among them, the number of positive samples is only a few dozen or less. To solve this issue, the two-step classification is introduced in RefineDet~\cite{DBLP:conf/cvpr/ZhangWBLL18}. It is a kind of cascade classification implemented through a two-step network architecture, in which the first step filters out most negative anchors using a preset threshold $\theta=0.99$ to reduce the search space for the subsequent step. Thus, the two-step classification can enhance the classification ability to reduce false positives.

However, it is unnecessary to perform two-step classification in all pyramid levels. Since the anchors tiled on the three higher levels (\ie, P5, P6, and P7) only account for $11.1\%$ and the associated features are much more adequate. Therefore, the classification task is relatively easy in these three higher pyramid levels. It is thus dispensable to apply two-step classification on the three higher pyramid levels, and if applied, it will lead to an increase in computation cost. In contrast, the three lower pyramid levels (\ie, P2, P3, and P4) have the vast majority of samples ($88.9\%$) and lack of adequate features. It is urgently needed for these low pyramid levels to do two-step classification in order to alleviate the class imbalance problem and reduce the search space for the subsequent classifier.

Therefore, our STC module selects C2, C3, C4, P2, P3, and P4 to perform two-step classification. As the statistical result shown in Figure~\ref{fig:sc}, the STC increases the positive/negative sample ratio by approximately $38$ times, from around $1$:$15441$ to $1$:$404$. In addition, we use the Focal loss in both two steps to make full use of samples. Unlike RefineDet~\cite{DBLP:conf/cvpr/ZhangWBLL18}, the RefineFace shares the same classification module in the two steps, since they have the same task to distinguish the face from the background. The experimental results of applying the two-step classification on each pyramid level are shown in Table~\ref{tab:stc_per_level}. Consistent with our analysis, the two-step classification on the three lower pyramid levels helps to improve performance, while on the three higher pyramid levels is ineffective.

The loss function for STC consists of two parts, \ie, the loss in the first step and the second step. For the first step, we calculate the focal loss for those samples selected to perform two-step classification. And for the second step, we just focus on those samples that remain after the first step filtering. With these definitions, we define the loss function:
\begin{small}
\begin{equation}
\begin{aligned}
{\cal L}_\text{STC}=\frac{1}{N_{\text{s}_3}} \sum_{i\in \Omega}{\cal L}_{\text{FL}}(p_i,l_i^\ast) + \frac{1}{N_{\text{s}_4}} \sum_{i\in \Phi}{\cal L}_{\text{FL}}(q_i,l_i^\ast),
\end{aligned}
\label{eqn2}
\end{equation}
\end{small}
{\flushleft where $p_i$ and $q_i$ are the predicted confidence of the anchor $i$ being a face in the first and second steps, $N_{\text{$s_3$}}$ and $N_{\text{$s_4$}}$ are the numbers of positive anchors in the first and second steps, $\Omega$ represents a collection of samples selected for two-step classification, $l_i^\ast$ and $\Phi$ are the same as defined in STR. The binary classification loss ${\cal L}_{\text{FL}}$ is the sigmoid focal loss over two classes (face {\em vs.} background).}

\subsection{Scale-aware Margin Loss}
To further improve the classification ability of our baseline, we borrow the idea of the margin-based loss function~\cite{DBLP:journals/corr/abs-1801-07698,DBLP:conf/icml/LiuWYY16,DBLP:journals/spl/WangCLL18,DBLP:conf/cvpr/WangWZJGZL018} from face recognition to face detection. The margin-based idea is a promising strategy in the face recognition task to improve the classification ability, which adds an extra margin to the classification loss to enhance the discrimination ability. Take the binary classification with sigmoid function for example. Supposed $x$ is the prediction value before the sigmoid output for a sample. Then the margin-based prediction is:

\begin{equation}
\begin{aligned}
y=sigmoid(x-m),
\end{aligned}
\label{eqn3}
\end{equation}
{\flushleft where $m$ is the margin added to $x$ and $y$ is the prediction probability. After that, $y$ is used in the classification loss, which can make the decision boundary more discriminative. Inspired by the margin-based loss function, we would like to add an extra margin to the sigmoid loss to improve the classification ability of face detectors.}

\begin{figure}[t]
\centering
\subfigure[Original decision surface]{
\label{fig:margin1} 
\includegraphics[width=0.8\linewidth]{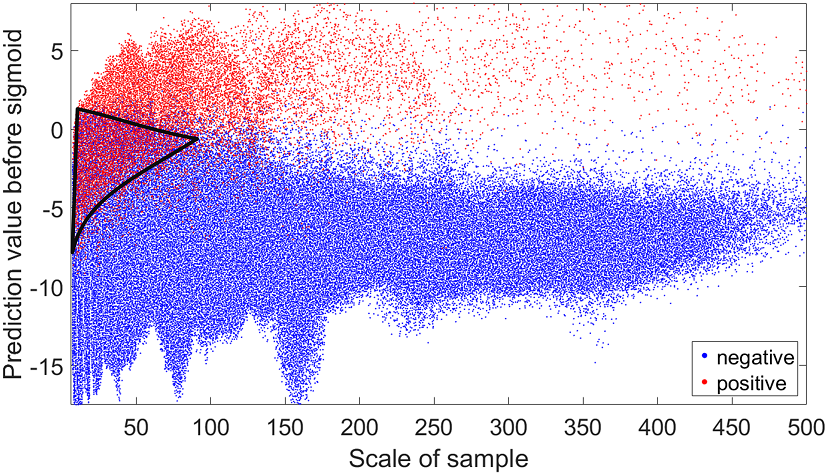}}
\subfigure[Decision surface with scale-aware margin loss]{
\label{fig:margin2} 
\includegraphics[width=0.8\linewidth]{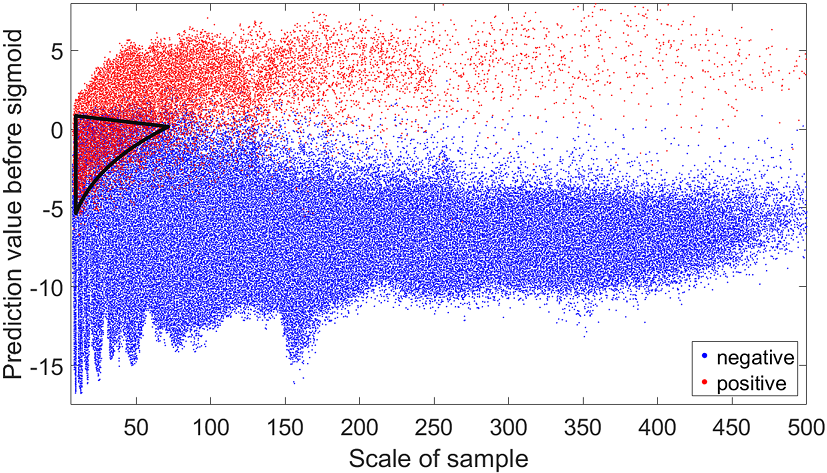}}
\caption{Visualization of classification ability. Red and blue points are positive and negative samples. The black curve enclosed area indicates the mixed region where face and background are indistinguishable.}
\label{fig:margin} 
\end{figure}

However, the constant margin used in face recognition is not suitable for face detection, because the binary classification task in face detection needs to classify samples of different scales. Specifically, for our anchor-based baseline, different scales of anchors are classified to detect various scales of faces. Small anchors have a large amount and are classified using coarser features, while large anchors have a small amount and are classified using detailed features. Thus, different scales of anchors have different degrees of classification difficulty and the classification of smaller anchors is more difficult. To elaborate on it, we visualize the decision boundary of our baseline detector at different scales on the WIDER FACE validation Hard subset as: feeding each image to the trained model and recording all the detections along with their scales (\ie, the square root of width$\times$height) and prediction scores before sigmoid, then showing all positive and negative samples in Figure~\ref{fig:margin1}. We can see that the decision boundary of the baseline detector becomes blurred as the scales become smaller. It means the classification ability becomes weaker as the scale of faces becomes smaller.

To handle this problem, we propose a scale-aware margin loss (SML) function, which adjusts the margin for each sample by its scale. The margin for each sample is given as:
\begin{equation}
\begin{aligned}
m={\alpha}/{\sqrt{wh}},
\end{aligned}
\label{eqn4}
\end{equation}
{\flushleft where $\alpha$ is a hyper-parameter to scale the margin, and $w$ and $h$ are the width and height of the sample. Larger faces have a more discriminative decision boundary so that they do not need too large margin; on the contrary, smaller faces have a blurred decision boundary so they need a larger margin to enhance the classification ability. After applying the scale-aware margin, the mixed region enclosed by the black curve becomes smaller as shown in Figure~\ref{fig:margin2}, suggesting the classification ability is enhanced for these small faces.}

\subsection{Feature Supervision Module}
\begin{figure}[t]
\centering
\includegraphics[width=0.45\textwidth]{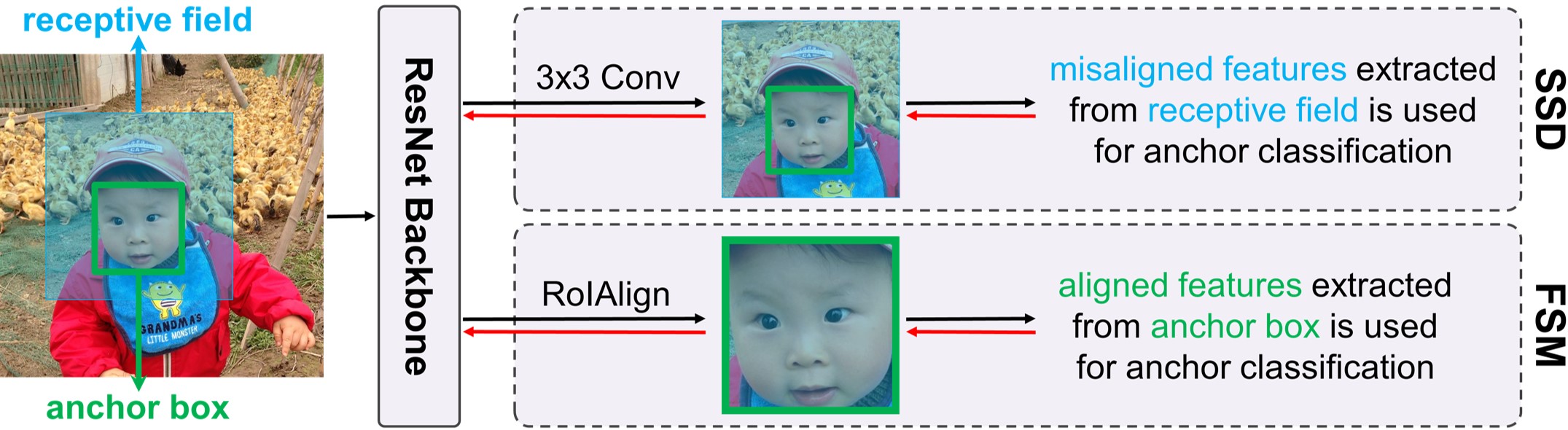}
\caption{Single-Shot Detectors (SSD) classify anchor using misaligned features extracted from receptive field, while FSM classifies anchor using aligned features extracted from anchor box. Black and red arrows mean forward and backward.}
\label{fig:fsm}
\end{figure}

Another reason for the limited classification ability of single-shot face detector is that the learned features in the backbone network are not sufficiently discriminative because of misalignment. Single-shot detectors perform face detection based on regular and dense anchors using fully convolutional network. As shown in Figure~\ref{fig:fsm}, the features used to classify the anchors in single-shot face detectors are misaligned, \ie, extracted from the corresponding receptive field and not tailored to the exact boundary of features in the anchor box. This shortcoming will make the learned features in the backbone not discriminative enough.

To solve this issue, we design a feature supervision module (FSM) to enable the single-shot backbone network to learn more discriminative features. This module is appended after the backbone network and classifies the anchors using aligned features extracted from anchor box. The backbone will be updated by the classification loss of FSM to learn more discriminative features. Specifically, this module first uses RoIAlign~\cite{DBLP:conf/iccv/HeGDG17} to extract the features at each detection, and then performs an extra binary classification based on the resultant features. This module has three characteristics: 1) we want to use it to enhance the classification ability, hence it only performs the binary classification; 2) it has a lightweight fully convolutional subnetwork with a relatively small loss, since it is an auxiliary module and should not over-dominate the training of the face detector; 3) it is not involved in the inference phase and will not introduce any additional overhead. With this additional supervision module, the backbone network is forced to learn more discriminative features for classification without any extra overhead during the inference phase.

To train the feature supervision module, we apply NMS with a threshold of $0.7$, add ground truth boxes, and distribute $512$ prediction proposals to the pyramid level from which they come to sample their RoI features. As shown in Figure~\ref{fig:framework}, the RoIAlign operation is performed at the assigned feature layers, yielding $5\times5$ resolution features, which are fed into three subsequent convolutional layers, a prediction convolutional layer and a global average pooling layer to classify between face and background.

\subsection{Receptive Field Enhancement}
At present, most detection networks utilize ResNet and VGGNet as the basic feature extraction module, while both of them possess square receptive fields. The singleness of the receptive field affects the detection of objects with different aspect ratios. This issue seems unimportant in face detection task, because the aspect ratio of face annotations is about $1$:$1$ in many datasets. Nevertheless, statistics show that the WIDER FACE training set has a considerable part of faces that have an aspect ratio of more than $2$ or less than $0.5$. Consequently, there is a mismatch between the receptive field of network and the aspect ratio of faces.

\begin{figure}[t!]
\centering
\subfigure[Structure]{
\label{fig:rfe1}
\includegraphics[width=0.45\linewidth]{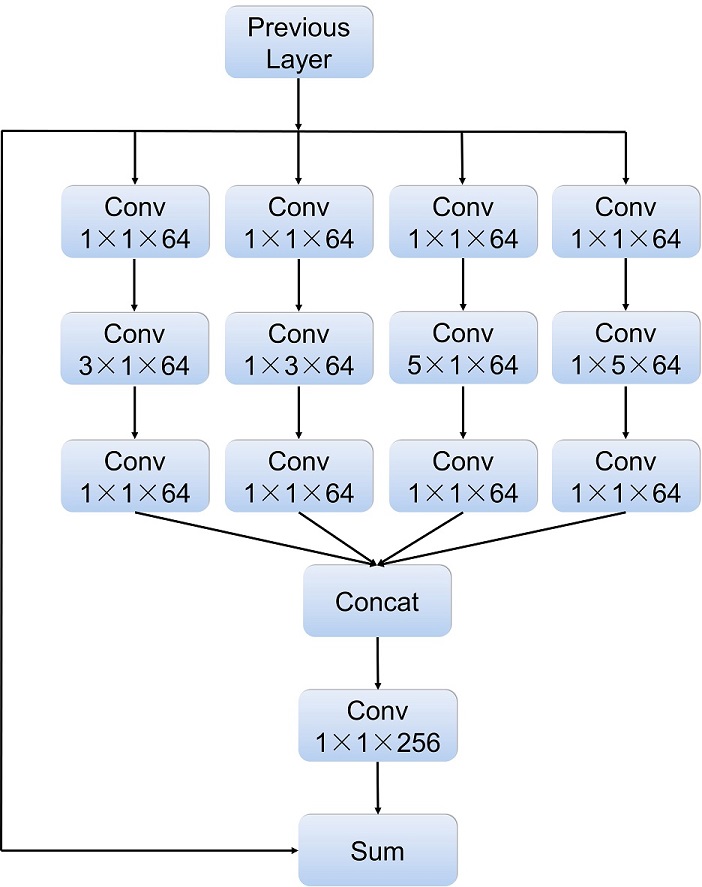}}
\subfigure[Illustration]{
\label{fig:rfe2}
\includegraphics[width=0.45\linewidth]{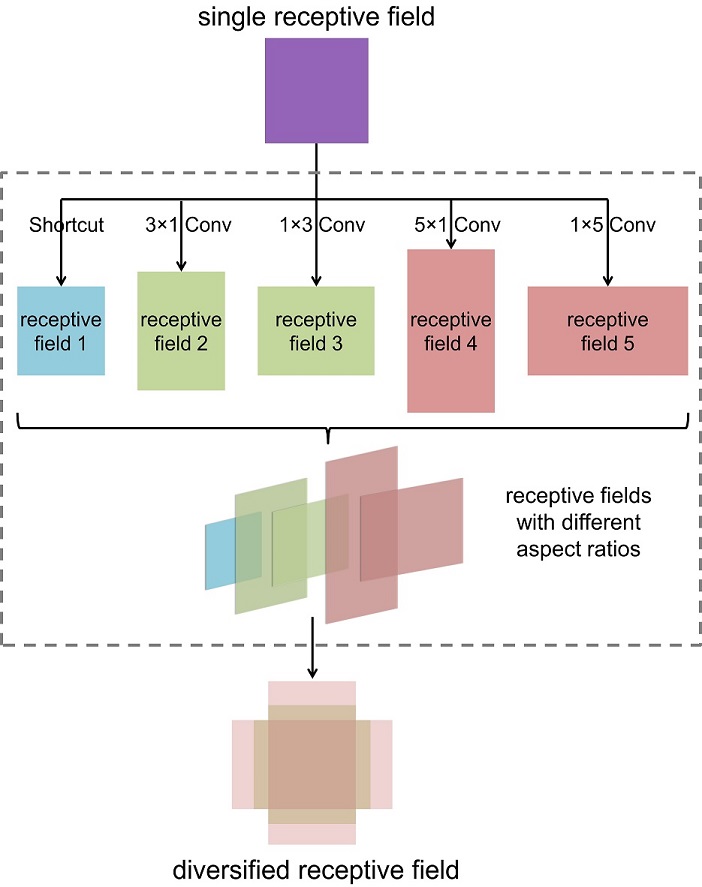}}
\caption{Structure and illustration of Receptive Field Enhancement (RFE).}
\label{fig:rfe}
\end{figure}

To address this issue, we propose a module named Receptive Field Enhancement (RFE) to diversify the receptive field of features before predicting classes and locations. In particular, RFE module replaces the middle two convolution layers in the class subnet and the box subnet of RetinaNet. The structure of RFE is shown in Figure~\ref{fig:rfe1}. Our RFE module adopts a four-branch structure, which is inspired by the Inception block \cite{DBLP:conf/cvpr/SzegedyLJSRAEVR15}. To be specific, first, we use a $1\times1$ convolution layer to decrease the channel number to one quarter of the previous layer. Second, we use $1\times k$ and $k\times 1$ ($k=3$ and $5$) convolution layer to provide rectangular receptive field. Through another $1\times1$ convolution layer, the feature maps from four branches are concatenated together. Additionally, we apply a shortcut path to retain the original receptive field from previous layer. As shown in Figure~\ref{fig:rfe2}, the RFE provides more diverse receptive fields that is helpful for detecting extreme-pose faces.

\subsection{Training and Inference}

{\flushleft \textbf{Data Augmentation.} }
To prevent over-fitting and construct a robust model, several data augmentation strategies are used to adapt to face variations, described as follows: 1) applying some photometric distortions to the training images; 2) expanding the images with a random factor in the interval $[1,2]$ by mean-padding; 3) cropping two square patches and randomly selecting one for training. One patch is with the size of the image's shorter side and the other one is with the size determined by multiplying a random number in the interval $[0.5,1.0]$ by the image's shorter side; 4) flipping the selected patch randomly and resizing it to $1024\times1024$ to get the final training sample.

{\flushleft \textbf{Anchor Design.} }
At each location of a detection layer, we associate two scales of anchors (corresponding to $2S$ and $2\sqrt{2}S$ in the original image, where $S$ is the downsampling factor of the detection layer) with one $1.25$ aspect ratio. So, there are $A=2$ anchors at each location of a detection layer, and it covers the scale of $8-362$ pixels in an input image.

{\noindent \textbf{Sample Matching.} }
During the training phase, the preset anchors and proposals need to be assigned as positive and negative samples for training. We assign anchors to ground-truth boxes using an intersection-over-union (IoU) threshold of $\theta_{p}$, and to background if their IoU is in $[0, \theta_{n})$. If an anchor is unassigned, which may happen with an IoU in $[\theta_{n}, \theta_{p})$, it is discarded during training. Empirically, we set $\theta_{n}=0.3$ and $\theta_{p}=0.7$ for the first step in STR and STC, $\theta_{n}=0.4$ and $\theta_{p}=0.5$ for the second step in STR and STC, and $\theta_{n}=0.4$ and $\theta_{p}=0.7$ for the feature supervision module.

{\flushleft \textbf{Loss Function.} }
We use a hybrid loss function to train the proposed model in an end-to-end fashion as ${\cal L} = {\cal L}_\text{STR} + {\cal L}_\text{STC} + {\cal L}_\text{FSM}$, where ${\cal L}_\text{STR}$ is the loss function for STR, ${\cal L}_\text{STC}$ is the loss function of STC, and ${\cal L}_\text{FSM}$ is the Focal loss for the binary classification in FAM.

{\flushleft \textbf{Optimization.} }
We use the ImageNet pretrained model to initialize the backbone network. All the parameters in the newly added convolution layers are initialized by the ``xavier" method. The stochastic gradient descent (SGD) algorithm is applied to fine-tune the RefineFace model with $0.9$ momentum, $0.0001$ weight decay and $32$ batch size. We use the warmup strategy to gradually ramp up the learning rate from $3.125\times10^{-4}$ to $1\times10^{-2}$ at the first $5$ epochs. After that, it switches to the regular learning rate schedule, \ie, dividing by $10$ at $100$ and $120$ epochs and ending at $130$ epochs. We use the PyTorch~\cite{paszke2017pytorch} library to implement the proposed RefineFace.

{\flushleft \textbf{Inference.} }
During the inference phase, the STC first filters the regularly tiled anchors on the selected pyramid levels with the negative confidence scores larger than the threshold $\theta=0.99$, and then STR adjusts the locations and sizes of selected anchors. After that, the second step takes over these refined anchors and outputs top $5,000$ detection results whose confidence scores are all higher than the threshold of $0.05$. Finally, we apply the non-maximum suppression (NMS) algorithm with Jaccard overlap of $0.4$ to generate top $750$ high confident detections per image as final results.

\section{Experiments}\label{4}

In this section, we conduct extensive experiments on WIDER FACE~\cite{DBLP:conf/cvpr/YangLLT16}, AFW~\cite{DBLP:conf/cvpr/ZhuR12}, PASCAL Face~\cite{DBLP:journals/ivc/YanZLL14}, FDDB~\cite{fddbTech} and MAFA~\cite{DBLP:conf/cvpr/GeLYL17} to verify the effectiveness of our RefineFace. Notably, our final model trained on WIDER FACE is directly evaluated on other datasets without finetuning.

\subsection{WIDER FACE Dataset}
It has $393,703$ annotated faces with variations in pose, scale, facial expression, occlusion and lighting condition in $32,203$ images. These images are split into three subsets: training ($40\%$), validation ($10\%$) and testing ($50\%$) sets. Each subset has three levels of difficulty: Easy, Medium and Hard based on the detection rate of EdgeBox~\cite{DBLP:conf/eccv/ZitnickD14}. All the models are trained on the training subset and tested on both the validation and testing subsets. Since the annotations of testing subsets are held-out, we submit the detection results to the collectors to report final evaluation results.

\subsubsection{Model Analysis}
To demonstrate the effectiveness of our proposed modules in RefineFace, each of them is added to our baseline to examine how it affects the final performance on the WIDER FACE dataset. All the experiments are based on ResNet-50 and use the same parameter settings for a fair comparison, except for specified changes. As listed in the first column of Table~\ref{tab:ablation}, our baseline detector based on RetinaNet achieves $95.1\%$ (Easy), $93.9\%$ (Medium) and $88.0\%$ (Hard) on the validation subset, which is better than most of face detectors on the WIDER FACE dataset. It can be considered as a strong single-shot face detector baseline and in the following, we will enhance its regression and classification ability to set a new state-of-the-art performance.

\begin{table}[t]
\centering
\setlength{\tabcolsep}{3.7pt}
\caption{Effectiveness of our proposed modules. Based on ResNet-50, all models are trained on WIDER FACE training subset and evaluated with AP ($\%$) on validation subset.}
\footnotesize{
\begin{tabular}{c|cccccccccc}
\toprule[1.5pt]
\multicolumn{1}{c|}{Module} & \multicolumn{10}{c}{RefineFace}\\
\hline
STR & & \Checkmark & & & & & \Checkmark & \Checkmark & \Checkmark & \Checkmark \\
STC & & & \Checkmark & & & & \Checkmark & \Checkmark & \Checkmark & \Checkmark \\
RFE & & & & \Checkmark & & & & \Checkmark & \Checkmark & \Checkmark \\
SML & & & & & \Checkmark & & & & \Checkmark & \Checkmark \\
FSM & & & & & & \Checkmark & & & & \Checkmark \\
\hline
{\em Easy}  & 95.1 & 95.9 & 95.3 & 95.5 & 95.5 & 95.8 & 96.1 & 96.4 & 96.6 &\textbf{96.9}\\
{\em Medium}& 93.9 & 94.8 & 94.4 & 94.3 & 94.6 & 94.5 & 95.0 & 95.3 & 95.6 &\textbf{95.9}\\
{\em Hard}  & 88.0 & 88.8 & 89.4 & 88.3 & 89.1 & 88.7 & 90.1 & 90.2 & 90.7 &\textbf{91.1}\\
\bottomrule[1.5pt]
\end{tabular}}
\label{tab:ablation}
\end{table}

{\flushleft \textbf{Selective Two-step Regression.} }
We add the STR module to our baseline detector to verify its effectiveness. As shown in Table~\ref{tab:ablation}, it produces much better results than the baseline, with $0.8\%$, $0.9\%$ and $0.8\%$ AP improvements on the Easy, Medium, and Hard subsets. Experimental results of applying two-step regression to each pyramid level (see Table~\ref{tab:str_per_level}) confirm our previous analysis. Inspired by the detection evaluation metric of MS COCO, we use $4$ IoU thresholds \{0.5, 0.6, 0.7, 0.8\} to compute the AP, so as to prove that the STR module can produce more accurate localization. As shown in Table~\ref{tab:aps}, the STR module produces consistently accurate detection results than the baseline method. The gap between the AP across all three subsets increases as the IoU threshold increases, which indicates that the STR module enhances the regression ability of our baseline and produces more accurate detections.

\begin{table}[!h]
\centering
\footnotesize
\caption{AP ($\%$) of the two-step regression applied to each pyramid level.}
\setlength{\tabcolsep}{7.5pt}
\begin{tabular}{c|c|cccccc}
\toprule[1.5pt]
STR & B & P2 & P3 & P4 & P5 & P6 & P7 \\
\hline
{\em Easy} & 95.1 & 94.8 & 94.3 & 94.8 & \bf 95.4 & \bf 95.7 & \bf 95.6 \\
{\em Medium} & 93.9 & 93.4 & 93.7 & 93.9 & \bf 94.2 & \bf 94.4 & \bf 94.6  \\
{\em Hard} & 88.0 & 87.5 & 87.7 & 87.0 & \bf 88.2 & \bf 88.2 & \bf 88.4 \\
\bottomrule[1.5pt]
\end{tabular}
\label{tab:str_per_level}
\end{table}

\begin{table}[!h]
\centering
\footnotesize
\caption{AP ($\%$) at different IoU thresholds on the WIDER FACE Hard subset.}
\setlength{\tabcolsep}{13.7pt}
\begin{tabular}{c|ccccc}
\toprule[1.5pt]
IoU & 0.5 & 0.6 & 0.7 & 0.8 \\
\hline
{RetinaNet} & 88.0 & 76.4 & 57.8 & 28.5\\
{RetinaNet+STR} & 88.8 & 83.4 & 66.5 & 38.2\\
\bottomrule[1.5pt]
\end{tabular}
\label{tab:aps}
\end{table}

{\flushleft \textbf{Selective Two-step Classification.} }
Experimental results of applying two-step classification to each pyramid level are shown in Table~\ref{tab:stc_per_level}, indicating that applying two-step classification to the low pyramid levels improves the performance, especially on tiny faces. Therefore, the STC module selectively applies the two-step classification on the low pyramid levels (\ie, P2, P3, and P4), since these levels are associated with lots of small anchors, which are the main source of false positives. As shown in Table~\ref{tab:ablation}, we find that after using the STC module, the AP scores of the detector are improved from $95.1\%$, $93.9\%$ and $88.0\%$ to $95.3\%$, $94.4\%$ and $89.4\%$ on the Easy, Medium and Hard subsets, respectively. In order to verify whether the improvements benefit from reducing the false positives, we count the number of false positives under different recall rates. As listed in Table~\ref{tab:fp_num}, our STC effectively reduces the false positives across different recall rates, demonstrating the effectiveness of the STC module. In addition, coupled with the STR module as listed in the seventh column of Table~\ref{tab:ablation}, the performance is further improved to $96.1\%$, $95.0\%$ and $90.1\%$ on the Easy, Medium and Hard subsets, respectively.

\begin{table}[!h]
\centering
\footnotesize
\caption{AP ($\%$) of the two-step classification applied to each pyramid level.}
\setlength{\tabcolsep}{7.7pt}
\begin{tabular}{c|c|cccccc}
\toprule[1.5pt]
STC & B & P2 & P3 & P4 & P5 & P6 & P7 \\
\hline
{\em Easy} & 95.1 & \bf 95.2 & \bf 95.2 & \bf 95.2 & 95.0 & 95.1 & 95.0 \\
{\em Medium} & 93.9 & \bf 94.2 & \bf 94.3 & \bf 94.1 & 93.9 & 93.7 & 93.9 \\
{\em Hard} & 88.0 & \bf 88.9 & \bf 88.7 & \bf 88.5 & 87.8 & 88.0 & 87.7 \\
\bottomrule[1.5pt]
\end{tabular}
\label{tab:stc_per_level}
\end{table}

\begin{table}[!h]
\centering
\footnotesize
\caption{Number of false positives at different recall rates.}
\setlength{\tabcolsep}{5pt}
\begin{tabular}{c|cccccc}
\toprule[1.5pt]
Recall ($\%$) & 10 & 30 & 50 & 80 & 90 & 95 \\
\hline
$\#$ FP of RetinaNet & 3 & 24 & 126 & 2,801 & 27,644 & 466,534\\
$\#$ FP of RetinaNet+STC & 1 & 20 & 101 & 2,124 & 13,163 & 103,586\\
\bottomrule[1.5pt]
\end{tabular}
\label{tab:fp_num}
\end{table}

{\flushleft \textbf{Receptive Field Enhancement.} }
The RFE is used to diversify the receptive fields of detection layers in order to capture faces with extreme poses. Comparing the detection results between first and fourth columns in Table~\ref{tab:ablation}, adding RFE to our baseline improves the AP performances by $0.4\%$, $0.4\%$ and $0.3\%$ for the Easy, Medium, and Hard subsets, respectively. Even though using RFE after STR and STC, it still consistently increases the AP scores in different subsets, \ie, from $96.1\%$ to $96.4\%$ for Easy, from $95.0\%$ to $95.3\%$ for Medium and from $90.1\%$ to $90.2\%$ for Hard. These improvements can be mainly attributed to the diverse receptive fields, which is useful to capture various pose faces for better detection accuracy. 

{\flushleft \textbf{Scale-aware Margin Loss.} }
We first only apply the scale-aware margin to the classification loss function of our baseline. Comparing the first and fifth columns in Table~\ref{tab:ablation}, we can observe that it improves the AP scores by $0.4\%$, $0.7\%$ and $1.1\%$ for the Easy, Medium, and Hard subsets respectively, benefiting from better discrimination between face and background, especially for small faces as shown in Figure~\ref{fig:margin2}. There is a hyper-parameter in Equation~\ref{eqn4} to scale the margin, we conduct several experiments to study its effect. We train the model with different $\alpha$ in $[3,7,11,15,19,23]$ on the WIDER FACE training set, then test on the validation set. As shown in Table~\ref{tab:hyperparam-alpha}, we observe that the proposed detector is relatively insensitive to the variations of $\alpha$. Too small value (\eg, $\alpha=3$) will make the margin not work, while too large (\eg, $\alpha=23$) would cause it difficult for training to be optimized. Thus, we choose $\alpha=15$ based on the validation performance in our experiments. Finally, we apply SML after STR, STC and RFE with a high starting point and the AP performances are still improved from $96.4\%$, $95.3\%$ and $90.2\%$ to $96.6\%$, $95.6\%$ and $90.7\%$ on the Easy, Medium and Hard subsets respectively. These results demonstrate its effectiveness.

\begin{table}[!h]
\centering
\footnotesize
\caption{AP ($\%$) of different $\alpha$ in the scale-aware margin loss.}
\setlength{\tabcolsep}{8pt}
\begin{tabular}{c|ccccccc}
\toprule[1.5pt]
$\alpha$ & 0 & 3 & 7 & 11 & 15 & 19 & 23 \\
\hline
{\em Easy} & 95.1 & 95.1 & 95.3 & 95.5 & \bf 95.5 & 95.3 & 95.0 \\
{\em Medium} & 93.9 & 93.9 & 94.3 & 94.5 & \bf 94.6 & 94.4 & 94.0 \\
{\em Hard} & 88.0 & 88.2 & 88.6 & 89.0 & \bf 89.1 & 89.1 & 88.9 \\
\bottomrule[1.5pt]
\end{tabular}
\label{tab:hyperparam-alpha}
\end{table}

{\flushleft \textbf{Feature Supervision Module.} }
To verify the effectiveness of the feature supervision module, we append it after our baseline and train the whole network end-to-end. As listed in the sixth column of Table~\ref{tab:ablation}, it boosts the AP results of our baseline by $0.7\%$, $0.6\%$ and $0.7\%$ for Easy, Medium and Hard subsets. These improvements come from the more discriminative features learned by the backbone network with the help of FSM. The output size of RoIAlign is a hyper-parameter and we conduct several experiments to select it. As shown in Table~\ref{tab:hyperparam-roi}, the moderate size $5\times5$ has the best results and too large or too small size will impair the performance. Besides, except the convolution (Conv) type, FAM can also be designed in the fully connected (FC) type. As shown in Table~\ref{tab:fam}, the Conv type achieves better performances with less parameters than the FC type, benefiting from that the Conv type shares parameters and retains spatial information. Notably, this module is only involved during training without any additional overhead during inference. Finally, FSM is added after other four modules, which still boost the AP performances from $96.6\%$, $95.6\%$, $90.7\%$ to $96.9\%$, $95.9\%$, $91.1\%$ on the Easy, Medium and Hard subsets respectively.

\begin{table}[!h]
\centering
\footnotesize
\caption{AP ($\%$) of different output sizes of RoIAlign in FAM.}
\setlength{\tabcolsep}{20pt}
\begin{tabular}{c|ccc}
\toprule[1.5pt]
Size & $3\times3$ & $5\times5$ & $7\times7$ \\
\hline
{\em Easy}   & 95.5 & \bf{95.8} & 95.7 \\
{\em Medium} & 94.3 & \bf{94.5} & 94.4 \\
{\em Hard}   & 88.6 & \bf{88.7} & 88.5 \\
\bottomrule[1.5pt]
\end{tabular}
\label{tab:hyperparam-roi}
\end{table}

\begin{table}[!h]
\centering
\footnotesize
\caption{AP ($\%$) of different design types of FAM.}
\setlength{\tabcolsep}{12.0pt}
\begin{tabular}{c|c|ccc}
\toprule[1.5pt]
Type & $\#$ parameter & {\em Easy} & {\em Medium} & {\em Hard} \\
\hline
None & 0 & 95.1 & 93.9 & 88.0 \\
FC   & 829,472 & 95.7 & 94.3 & 88.5 \\
Conv & 387,360 & \bf{95.8} & \bf{94.5} & \bf{88.7} \\
\bottomrule[1.5pt]
\end{tabular}
\label{tab:fam}
\end{table}

\begin{figure*}[!ht]
\centering
\subfigure[Val: Easy]{
\label{fig:ve} 
\includegraphics[width=0.315\linewidth]{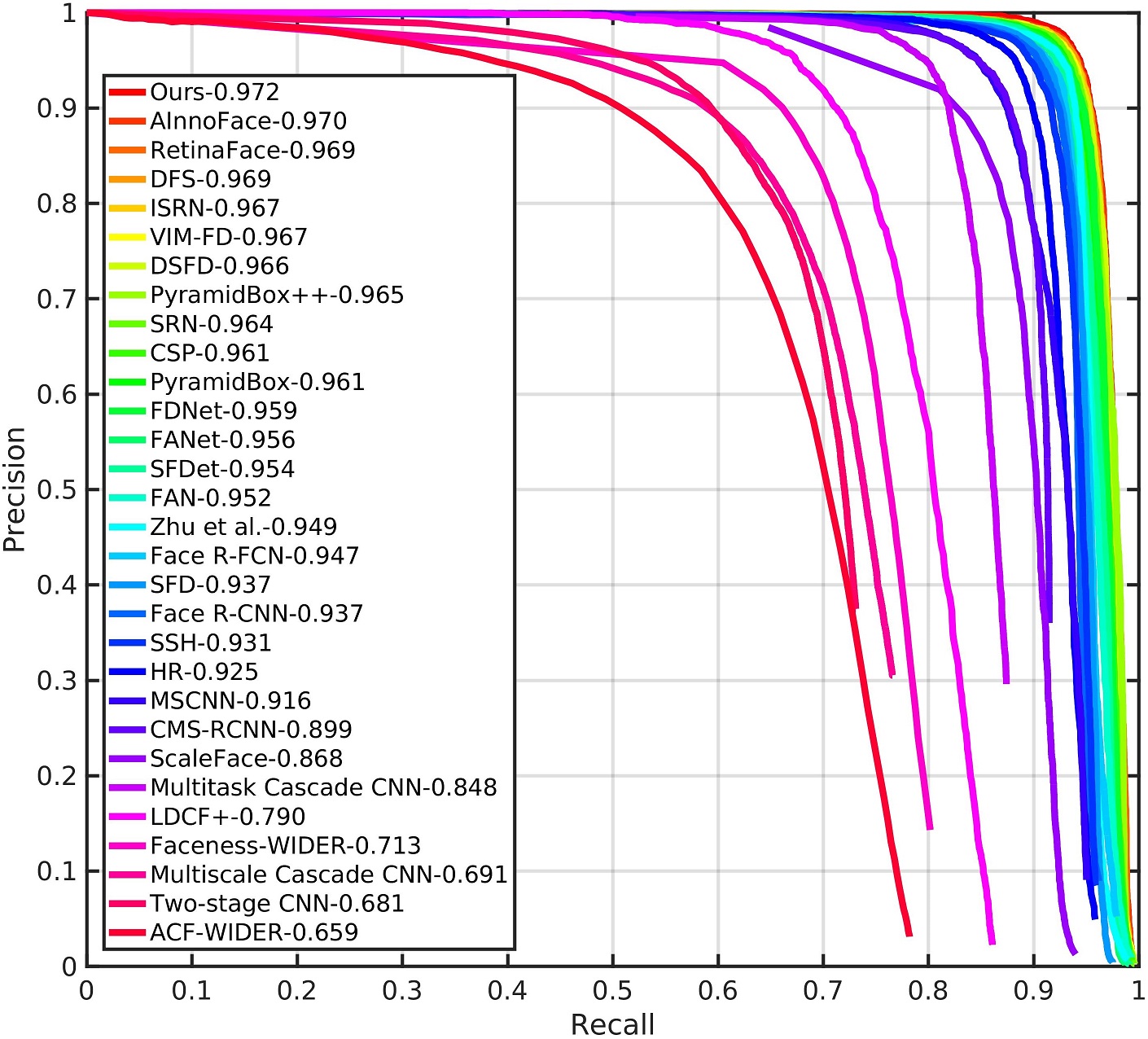}}
\subfigure[Val: Medium]{
\label{fig:vm} 
\includegraphics[width=0.315\linewidth]{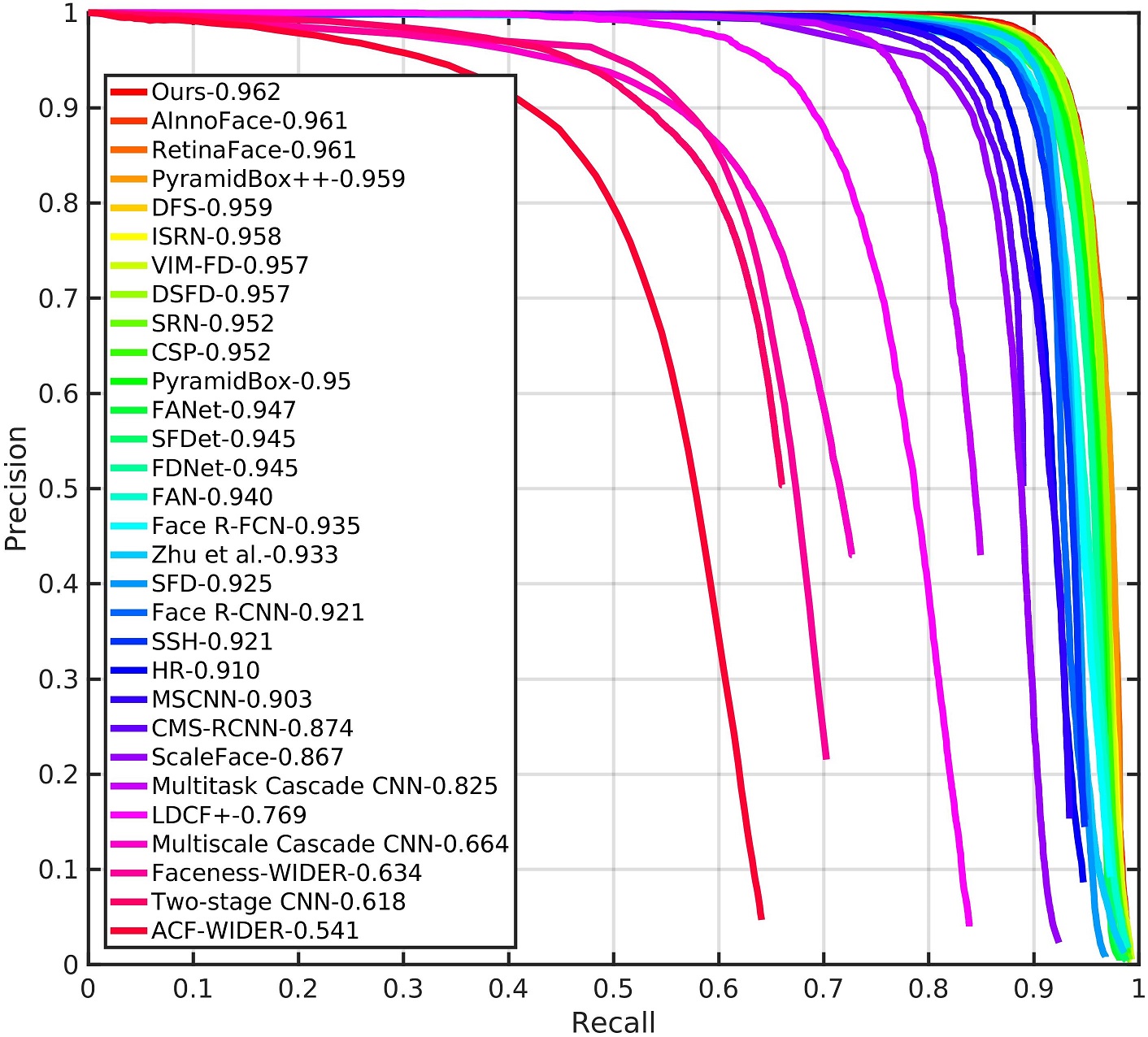}}
\subfigure[Val: Hard]{
\label{fig:vh} 
\includegraphics[width=0.315\linewidth]{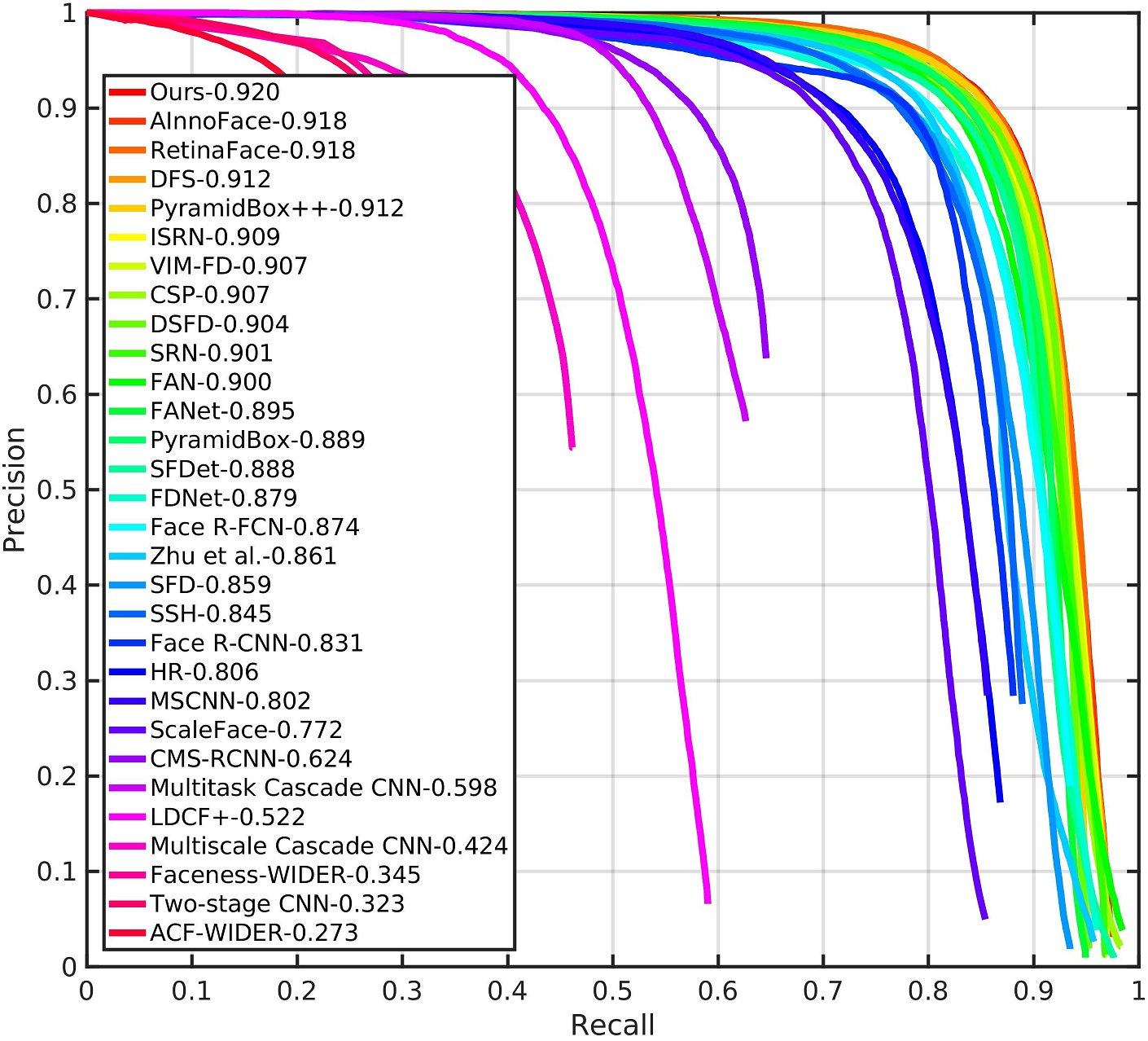}}
\subfigure[Test: Easy]{
\label{fig:te} 
\includegraphics[width=0.315\linewidth]{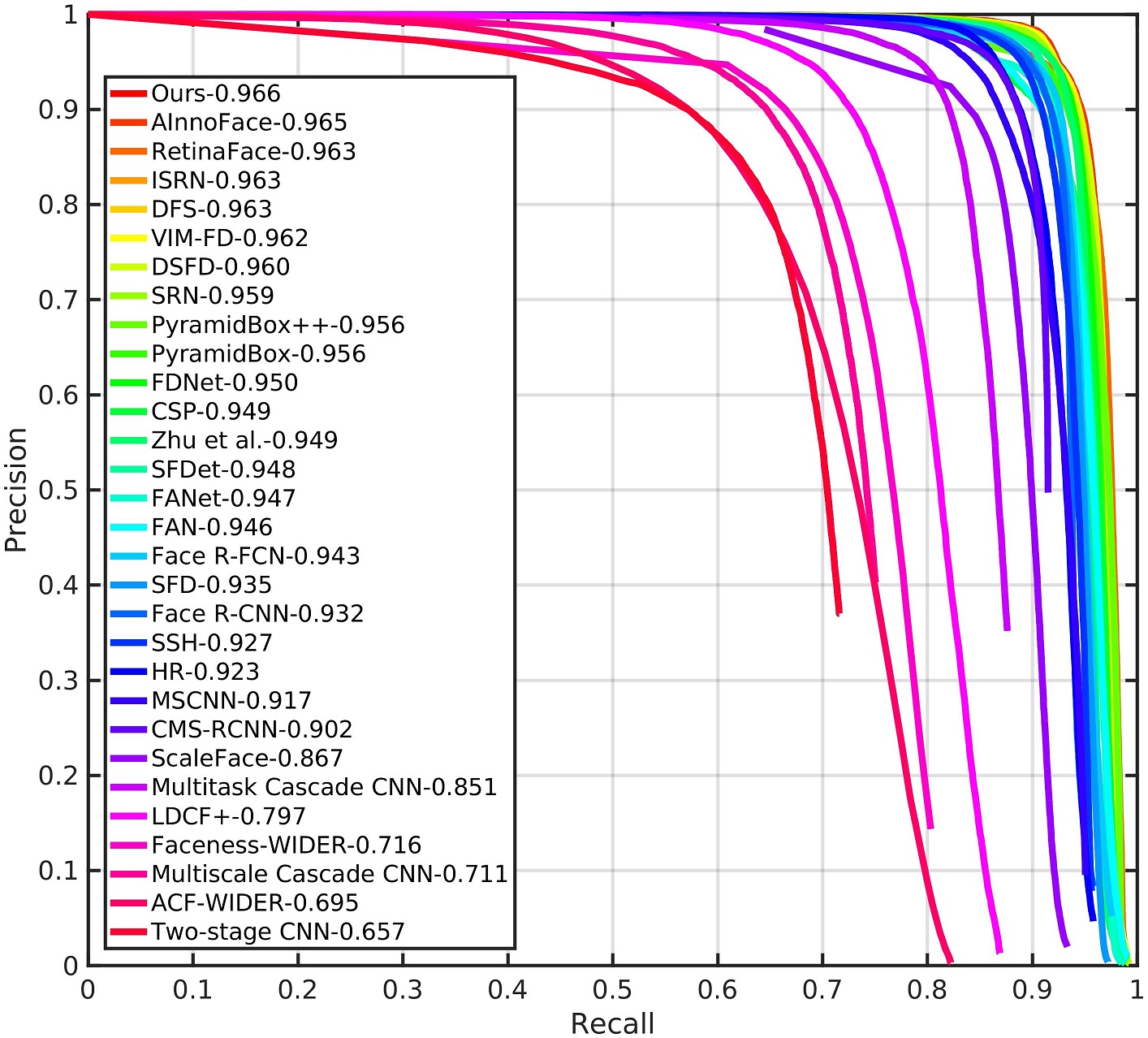}}
\subfigure[Test: Medium]{
\label{fig:tm} 
\includegraphics[width=0.315\linewidth]{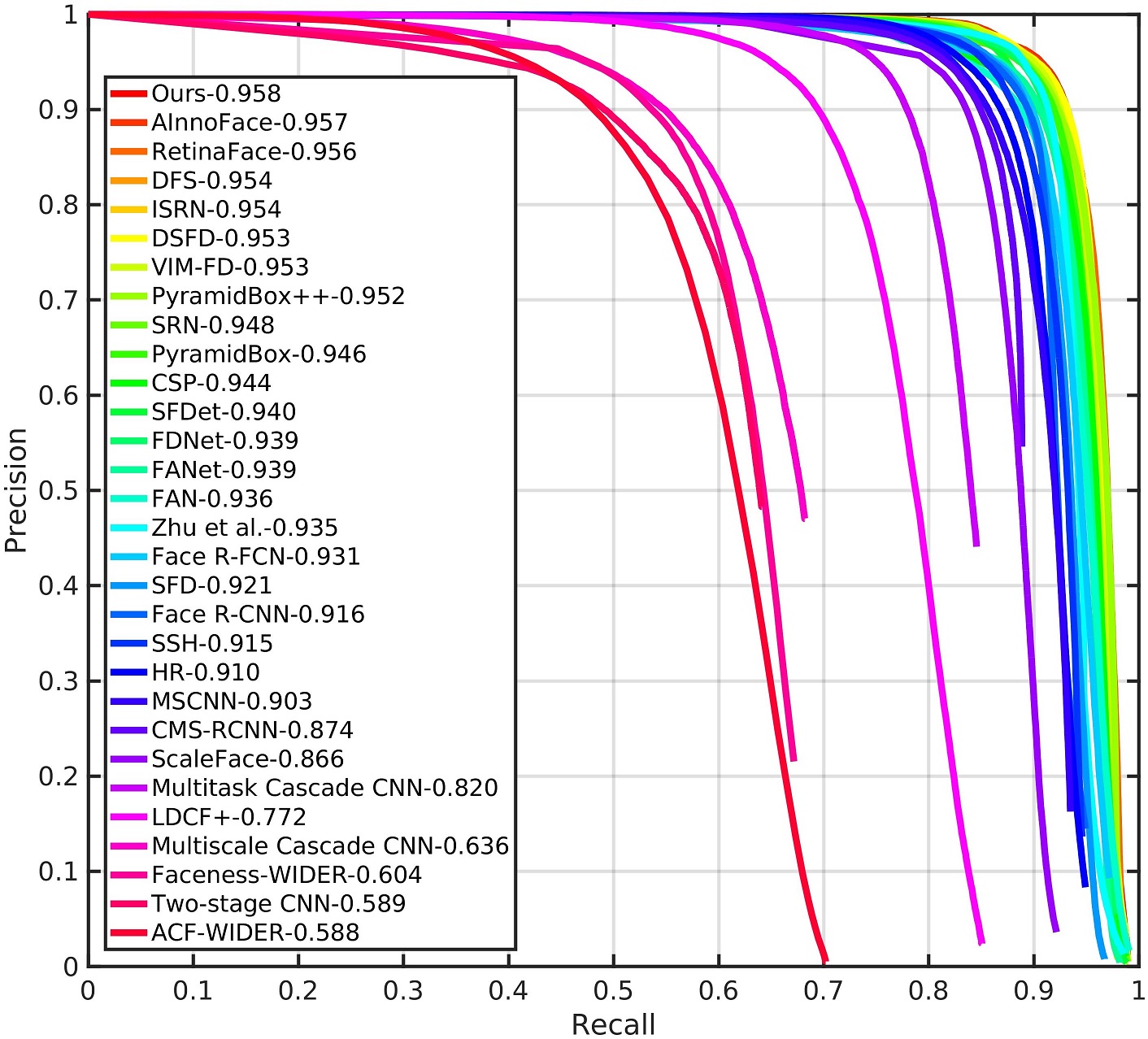}}
\subfigure[Test: Hard]{
\label{fig:th} 
\includegraphics[width=0.315\linewidth]{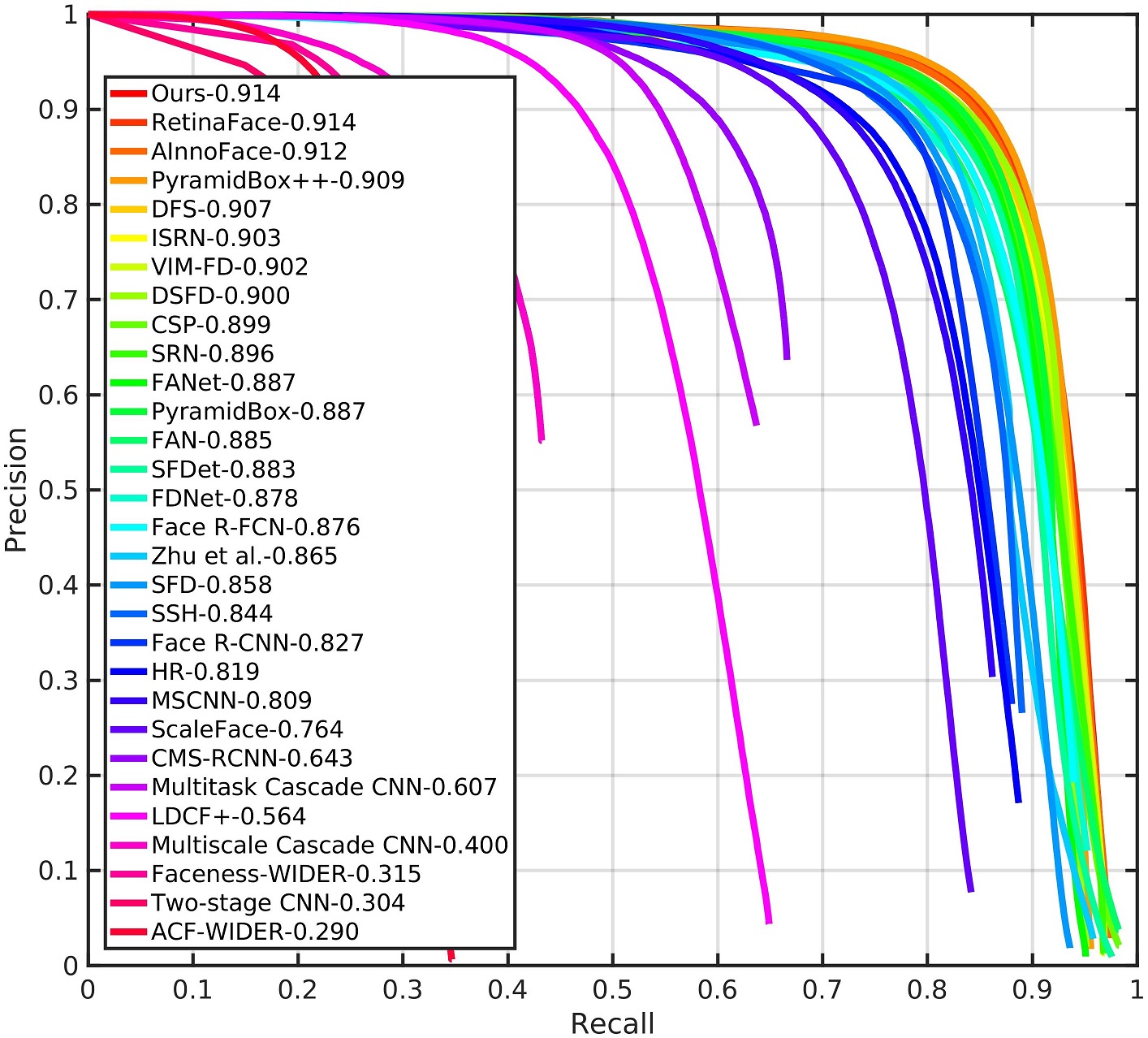}}
\caption{Precision-recall curves on WIDER FACE validation and testing subsets.}
\label{fig:wider-face} 
\end{figure*}

\subsubsection{Tradeoff Analysis}
All the experiments in last subsection are based on ResNet-50 and in this section we will analysis the trade-off between speed and accuracy of different backbone networks. The inference time of RefineFace are measured on a single NVIDIA GTX 1080-Ti with CUDA 9.0 and cuDNN v7.0. Frame-per-second (FPS) and millisecond (ms) are used to compare the speed. As listed in Table~\ref{tab:tradeoff}, our best model with ResNet-152 backbone can run at $17.7$ FPS ($56.6$ ms) for the VGA-resolution image and ResNet-101 achieves promising performances with near real-time speed. We also utilize more lightweight backbones to balance between speed and accuracy. Based on more efficient backbones, the proposed model can achieve real-time speed ($\geq25$ FPS) for the VGA-resolution image, \ie, $28.5$ FPS ($35.1$ ms) for ResNet-50 and $37.3$ FPS ($26.8$ ms) for ResNet-18. Among these backbones, ResNet-152 has the best performance with the slowest speed, ResNet-18 reaches the fastest speed with the lowest AP, while ResNet-50 has the best balance between speed and accuracy, which is why most of models~\cite{DBLP:conf/aaai/abs-1809-02693,DBLP:journals/corr/abs-1811-08557,DBLP:journals/corr/abs-1711-07246,DBLP:journals/corr/abs-1901-06651,DBLP:journals/corr/abs-1901-02350} use ResNet-50 as the backbone network to build the face detectors.

\begin{table}[h]
\centering
\footnotesize
\setlength{\tabcolsep}{9.5pt}
\caption{Tradeoff analysis between speed and accuracy. Speed is measured with a VGA-resolution ($640\times480$) input image. All batch normalization (BN) layers are merged into the convolution layers during inference.}
\begin{tabular}{c|c|c|ccc}
\toprule[1.5pt]
\multirow{2}{*}{Backbone} & \multicolumn{2}{c|}{Speed} & \multicolumn{3}{c}{Accuracy ($\%$)} \\
\cline{2-6}
 & FPS & ms & Easy & Medium & Hard \\
\hline
ResNet-18  & 37.3 & 26.8 & 96.3 & 95.1 & 90.2 \\
ResNet-50  & 28.5 & 35.1 & 96.9 & 95.9 & 91.1 \\
ResNet-101 & 22.8 & 43.9 & 97.1 & 96.1 & 91.6 \\
ResNet-152 & 17.7 & 56.6 & 97.2 & 96.2 & 92.0 \\
\bottomrule[1.5pt]
\end{tabular}
\label{tab:tradeoff}
\end{table}

\subsubsection{Performance Analysis}
As shown in Figure~\ref{fig:wider-face}, we compare RefineFace with $29$ state-of-the-art face detection methods~\cite{DBLP:conf/eccv/CaiFFV16,DBLP:conf/aaai/abs-1809-02693,DBLP:conf/cvpr/HuR17,li2018dsfd,DBLP:conf/iccv/NajibiSCD17,DBLP:conf/icpr/Ohn-BarT16a,tang2018pyramidbox,DBLP:journals/corr/abs-1811-08557,DBLP:journals/corr/WangLJW17,DBLP:journals/corr/abs-1711-07246,DBLP:journals/corr/abs-1709-05256,DBLP:conf/icb/YangYLL14,DBLP:conf/iccv/YangLLT15,DBLP:conf/cvpr/YangLLT16,DBLP:journals/corr/YangXLT17,DBLP:journals/corr/abs-1802-02142,DBLP:journals/corr/abs-1712-00721,DBLP:journals/spl/ZhangZLQ16,DBLP:conf/iccv/abs-1708-05237,zhu2018seeing,DBLP:journals/corr/ZhuZLS16,DBLP:journals/corr/abs-1905-01585,DBLP:journals/corr/abs-1904-02948,DBLP:journals/corr/abs-1901-06651,DBLP:journals/corr/abs-1901-02350,DBLP:journals/ijcv/ZhangWSLLL19,DBLP:journals/corr/abs-1905-00641,DBLP:journals/corr/abs-1904-00386} on both the validation and testing subsets. The proposed RefineFace achieves the best AP performance in all subsets of both validation and testing sets, \ie, $97.2\%$ (Easy), $96.2\%$ (Medium) and $92.0\%$ (Hard) for validation set, and $96.6\%$ (Easy), $95.8\%$ (Medium) and $91.4\%$ (Hard) for testing set. It outperforms all compared state-of-the-art methods based on the average precision (AP) across the three subsets, demonstrating the superiority of the proposed face detector. Notably, among all the published methods~\cite{DBLP:conf/eccv/CaiFFV16,DBLP:conf/aaai/abs-1809-02693,DBLP:conf/cvpr/HuR17,li2018dsfd,DBLP:conf/iccv/NajibiSCD17,DBLP:conf/icpr/Ohn-BarT16a,tang2018pyramidbox,DBLP:conf/icb/YangYLL14,DBLP:conf/iccv/YangLLT15,DBLP:conf/cvpr/YangLLT16,DBLP:journals/spl/ZhangZLQ16,DBLP:conf/iccv/abs-1708-05237,zhu2018seeing,DBLP:journals/corr/abs-1904-02948,DBLP:journals/ijcv/ZhangWSLLL19}, our method outperforms the previous best method DSFD~\cite{li2018dsfd} by a large margin. Although there are several unpublished technical reports have very promising performances, some of them~\cite{DBLP:journals/corr/abs-1901-06651,DBLP:journals/corr/abs-1905-00641} use additional data, some of them~\cite{DBLP:journals/corr/abs-1811-08557,DBLP:journals/corr/abs-1905-01585,DBLP:journals/corr/abs-1901-02350,DBLP:journals/corr/abs-1904-00386} apply some existing time-consuming tricks including segmentation, attention and context. In contrast, the proposed method presents five new modules with a small amount of additional overhead and the five modules are complementary to existing methods. 

\subsection{AFW dataset}
\begin{figure}[h]
\centering
\includegraphics[width=0.45\textwidth]{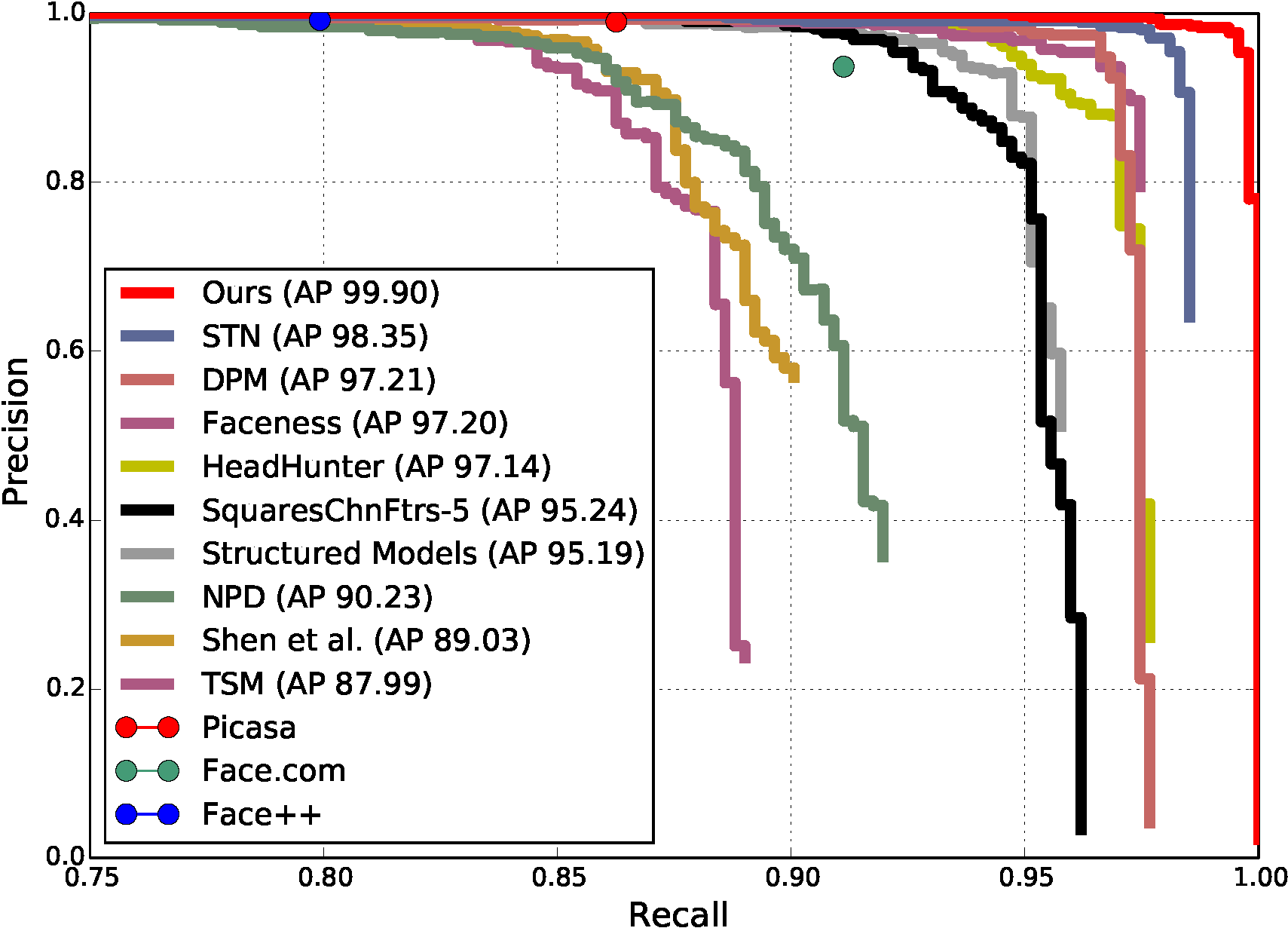}
\caption{Precision-recall curves on the AFW dataset.}
\label{fig:afw}
\end{figure}

It contains $473$ labeled faces in $205$ images, which are collected from Flickr and have cluttered backgrounds with large variations in both face viewpoints and appearances (\eg, ages, sunglasses, make-ups, skin colors, expressions, etc.). As shown by the precision-recall curves in Figure~\ref{fig:afw}, we compare RefineFace against three commercial face detectors (\ie, Face.com, Face++ and Picasa) and nine state-of-the-art methods~\cite{DBLP:conf/eccv/ChenHW016,DBLP:journals/pami/LiaoJL16,DBLP:conf/eccv/MathiasBPG14,DBLP:conf/cvpr/ShenLBW13,DBLP:journals/ivc/YanZLL14,DBLP:conf/iccv/YangLLT15,DBLP:conf/cvpr/ZhuR12}. The proposed method improves the AP score of state-of-the-art results by $1.55\%$ compared with the second best method STN~\cite{DBLP:conf/eccv/ChenHW016}.

\subsection{PASCAL Face datatset}
\begin{figure}[h]
\centering
\includegraphics[width=0.45\textwidth]{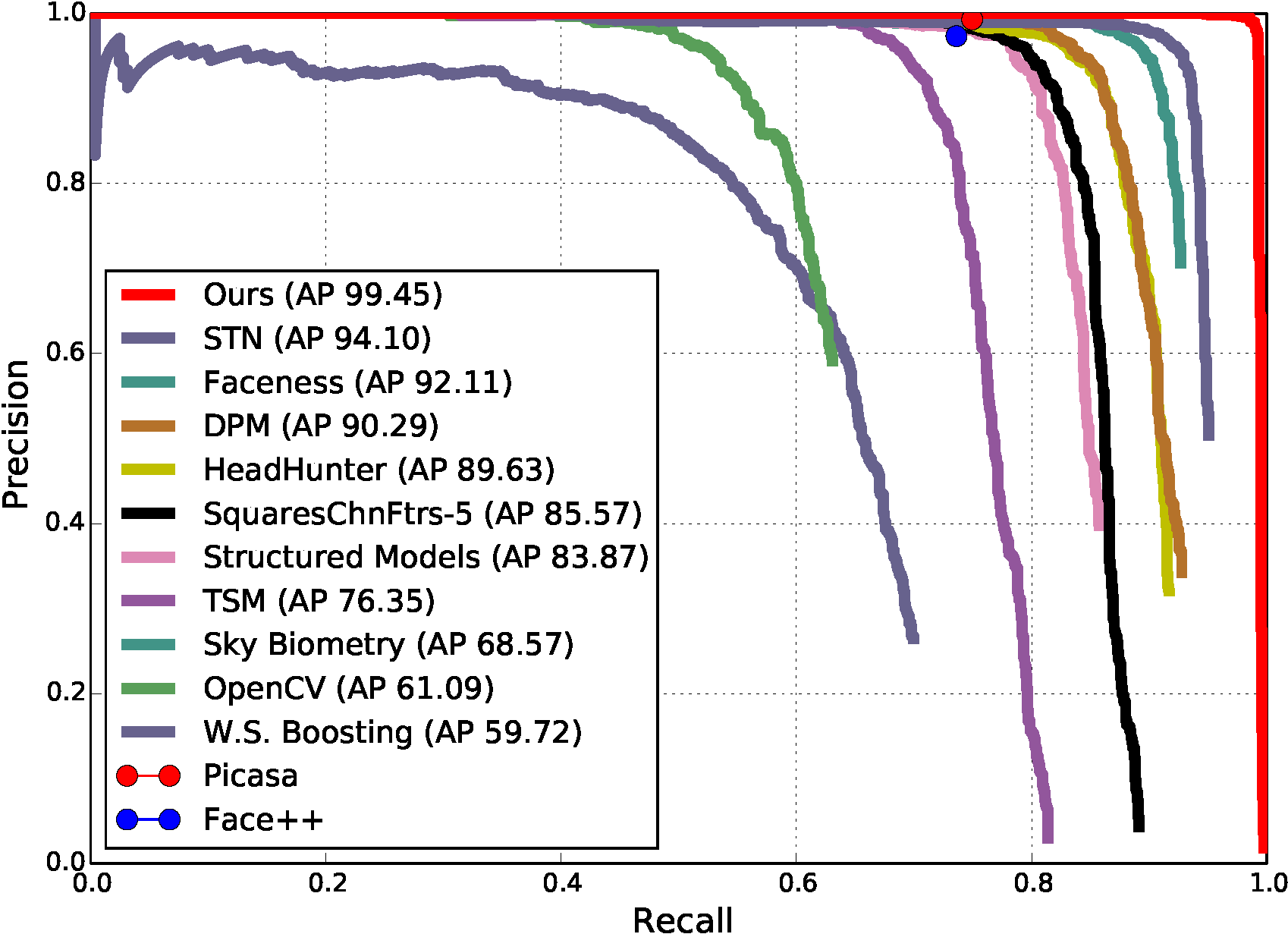}
\caption{Precision-recall curves on the PASCAL Face dataset.}
\label{fig:PASCAL}
\end{figure}

It is collected from PASCAL person layout test subset and consists of $1,335$ labeled faces in $851$ images with large face appearance and pose variations. Figure~\ref{fig:PASCAL} shows the precision-recall curves of the proposed RefineFace compared with $9$ state-of-the-art methods~\cite{DBLP:conf/eccv/ChenHW016,DBLP:conf/bmvc/KalalMM08,DBLP:conf/eccv/MathiasBPG14,DBLP:journals/ivc/YanZLL14,DBLP:conf/iccv/YangLLT15,DBLP:conf/cvpr/ZhuR12} and $3$ commercial face detectors (\ie, SkyBiometry, Face++ and Picasa). The RefineFace model outperforms the state-of-the-art methods with the top AP score ($99.45\%$).

\subsection{FDDB dataset}
\begin{figure}[h]
\centering
\includegraphics[width=0.45\textwidth]{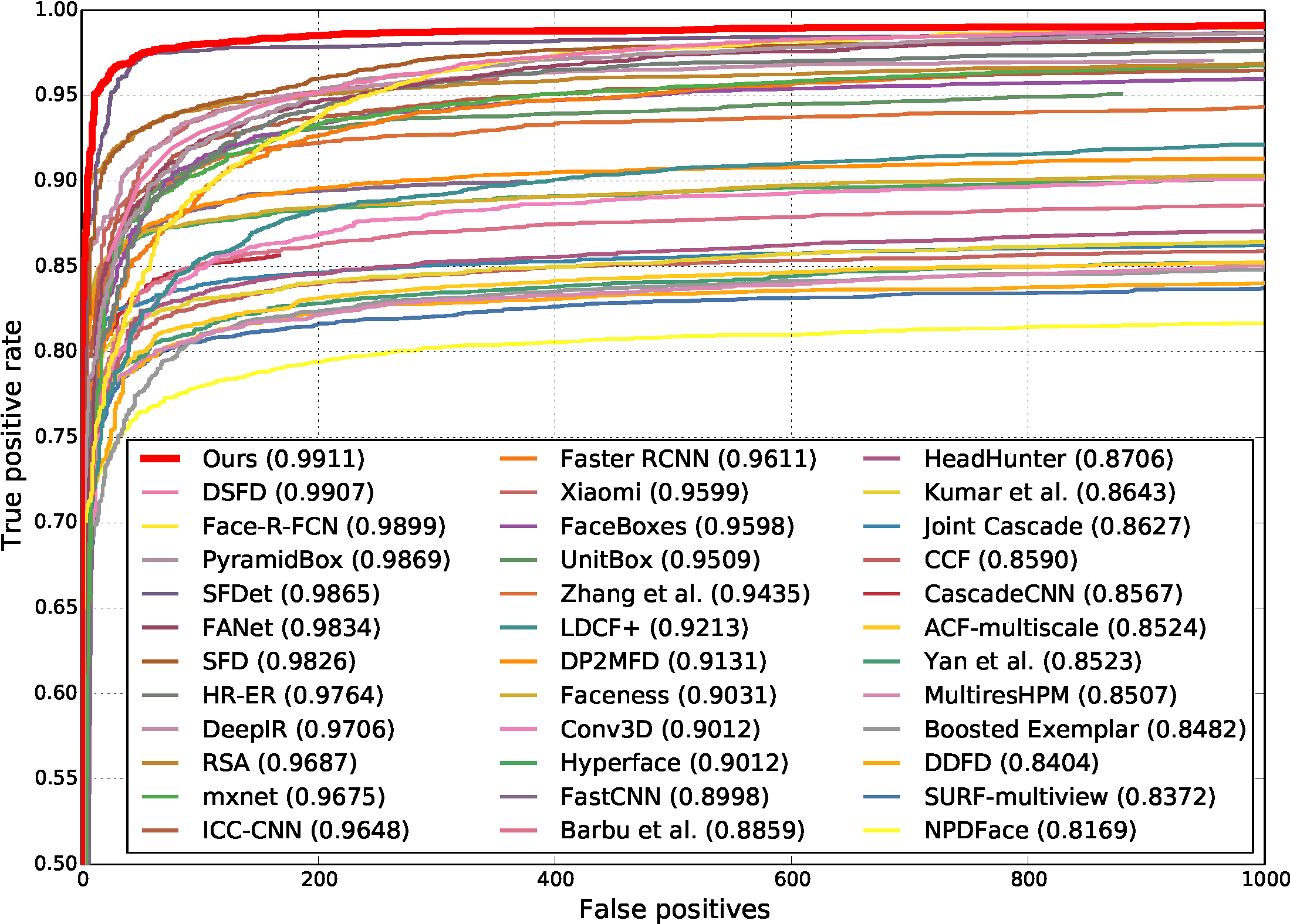}
\caption{Receiver operating characteristic (ROC) curves with discrete scores on the FDDB dataset. The number in the legend is the true positive rate (TPR) at the false positives (FP) equals to $1,000$.}
\label{fig:fddb}
\end{figure}

It has $5,171$ faces annotated in $2,845$ images with a wide range of challenges including low image resolutions, severe occlusions and difficult poses. Since there are lots of unlabelled faces on FDDB, which results in many false positive faces with high scores. Hence, we use the new annotations~\cite{DBLP:journals/ijcv/ZhangWSLLL19} to evaluate the proposed detector and compare it with several state-of-the-art methods~\cite{barbu2014face,DBLP:conf/mir/FarfadeSL15,DBLP:journals/corr/GhiasiF15,DBLP:conf/cvpr/HuR17,DBLP:conf/fgr/JiangL17,DBLP:conf/iccv/KumarNJ15,DBLP:conf/cvpr/LiLBSH14,li2018dsfd,DBLP:conf/cvpr/Li013,DBLP:conf/eccv/LiSWW16,DBLP:journals/pami/LiaoJL16,DBLP:conf/iccv/LiuLYWWT17,DBLP:conf/icpr/Ohn-BarT16a,DBLP:conf/btas/RanjanPC15,DBLP:journals/pami/RanjanPC19,DBLP:journals/ijon/SunWH18,tang2018pyramidbox,DBLP:conf/inns/Triantafyllidou16,DBLP:journals/corr/WanCZZW16,DBLP:journals/corr/abs-1709-05256,DBLP:conf/iccv/YangLLT15,DBLP:conf/mm/YuJWCH16,DBLP:journals/corr/abs-1712-00721,DBLP:journals/spl/ZhangZLQ16,DBLP:conf/iccv/zhang2017detecting,DBLP:conf/ijcb/abs-1708-05234,DBLP:conf/iccv/abs-1708-05237} in Figure~\ref{fig:fddb}. The proposed face detector achieves $99.11\%$ true positive rate when the number of false positives is equal to $1,000$, setting a new state-of-the-art result. It indicates the superior performance of RefineFace in presence of various scales, large appearance variations, heavy occlusions and severe blur degradations in unconstrained scenarios. 

\subsection{MAFA Dataset}
It is a benchmark dataset for detecting occluded faces and contains $35,806$ masked faces in $30,811$ images collected from Internet. This dataset covers $60$ cases of occluded faces in daily scenarios with three degrees of occlusions, four types of masks and five face orientations. In this dataset, if the face is severely blurred, deformed, or smaller than $32$ pixels, it is ignored. The left are divided into masked faces and unmasked faces. Therefore, there are three subsets on the MAFA dataset: (1) the whole subset (containing $6,354$ masked faces, $996$ unmasked faces and $2,683$ ignored faces), (2) the masked subset (including masked faces), and (3) the unignored subset (consisting of masked faces and unmasked faces). We directly evaluate the model trained on WIDER FACE with all three subsets and report the average precision (AP) scores against eight state-of-the-art methods in Table~\ref{tab:mafa}. Our method sets a new state-of-the-art result on all three subsets, \ie, $83.9\%$ on the whole subset, $96.2\%$ on the masked subset and $95.7\%$ on the unignored subset. These results demonstrate that the proposed method is robust to various occlusions in our daily scenarios.

\begin{table}[h]
\centering
\footnotesize
\setlength{\tabcolsep}{8pt}
\caption{AP ($\%$) on the MAFA testing set.}
\begin{tabular}{c|c|c|c}
\toprule[1.5pt]
Methods & Whole set & Masked set & Unignored set \\
\hline
TSM~\cite{DBLP:conf/cvpr/ZhuR12} & - & - & 41.6\\
HeadHunter~\cite{DBLP:conf/eccv/MathiasBPG14} & - & - & 50.9\\
HPM~\cite{DBLP:journals/corr/GhiasiF15} & - & - & 60.0\\
MTCNN~\cite{DBLP:journals/spl/ZhangZLQ16} & - & - & 60.8\\
LLE-CNNs~\cite{DBLP:conf/cvpr/GeLYL17} & - & - & 76.4\\
FAN~\cite{DBLP:journals/corr/abs-1711-07246} & - & 76.5 & 88.3\\
AOFD~\cite{DBLP:journals/corr/abs-1709-05188} & 81.3 & 83.5 & 91.9\\
SFDet~\cite{DBLP:journals/ijcv/ZhangWSLLL19} & 81.5 & 94.0 & 93.7\\
\hline
\textbf{Ours} & \textbf{83.9} & \textbf{96.2} & \textbf{95.7}\\
\bottomrule[1.5pt]
\end{tabular}
\label{tab:mafa}
\end{table}

\section{Conclusion} \label{5}
This paper proposes a state-of-the-art single-shot face detector by enhancing the regression and classification ability. On the one hand, boosting the regression ability can improve the location accuracy and reduce the LOC error, for this purpose we design the STR to coarsely adjust the locations and sizes of anchors from high level detection layers to provide better initialization for the subsequent regressor. On the other hand, enhancing the classification ability can improve the recall efficiency and reduce the CLS error, to this end we first use the STC to filter out most simple negatives from low level detection layers to reduce the search space for the subsequent classifier, then apply the SML to better distinguish faces from background at various scales and the FSM to let the backbone network learn more discriminative features for classification. In addition, we introduce the RFE to provide more diverse receptive field to better capture faces in some extreme poses. Experiments are conducted on most of challenging face detection datasets to demonstrate the effectiveness of RefineFace. In the future, we plan to design a lightweight architecture with the help of automatic machine learning (AutoML) methods to make RefineFace run in real-time not only on the GPU device but also on the CPU device.

\bibliographystyle{IEEEtran}
\bibliography{reference}

\end{document}